%% file: main.tex
\documentclass[runningheads]{llncs}

\usepackage[review,year=2026,ID=14130]{eccv}
\usepackage{eccvabbrv}

\usepackage{graphicx}
\usepackage{subcaption}
\usepackage[table]{xcolor}
\usepackage{booktabs}
\usepackage{xurl}
\usepackage{amsmath}
\usepackage{amssymb}
\usepackage{mathtools}
\usepackage{multirow}
\usepackage{enumitem}
\usepackage[accsupp]{axessibility} % accessibility helpers
\usepackage{pifont} % for \ding
\usepackage{float}            % [H] placement for figures
\usepackage{placeins}         % \FloatBarrier for local float control
\usepackage[most]{tcolorbox} % needed by supplemental prompt box
\usepackage[pagebackref,breaklinks,colorlinks,citecolor=eccvblue]{hyperref}

\input{math_commands.tex}

\usepackage[textsize=tiny]{todonotes}

\AtBeginDocument{\nolinenumbers}

\begin{document}

\title{LatentGeo: Learnable Auxiliary Constructions in Latent Space for Multimodal Geometric Reasoning}
\titlerunning{LatentGeo: Learnable Auxiliary Constructions}

\author{\textbf{Haiying Xu}\inst{1,2}\textsuperscript{*} \and
\textbf{Zihan Wang}\inst{1,3}\textsuperscript{*} \and
\textbf{Song Dai}\inst{1}\textsuperscript{*} \and
\textbf{Zhengxuan Zhang}\inst{1} \\[0.2em]
\textbf{Kairan Dou}\inst{2} \and
\textbf{Xuming Hu}\inst{1}\textsuperscript{\textdagger}}

% Short author list for running head
\authorrunning{Xu et al.}

\institute{$^{1}$ The Hong Kong University of Science and Technology (Guangzhou)\\
$^{2}$ Nankai University \quad $^{3}$ Communication University of China}

\makeatletter
\@ifundefined{maketitleold}{%
  \newcommand{\eccvarxivmaketitle}{\maketitle}%
}{%
  \newcommand{\eccvarxivmaketitle}{\maketitleold}%
}
\makeatother

\eccvarxivmaketitle

\vspace{-0.4cm}
\begingroup
\makeatletter
\renewcommand\thefootnote{\@fnsymbol\c@footnote}
\footnotetext[1]{Equal contribution. Work done during Haiying Xu and Zihan Wang's internship at HKUST(GZ). \raisebox{-0.18em}{\includegraphics[height=1.1em]{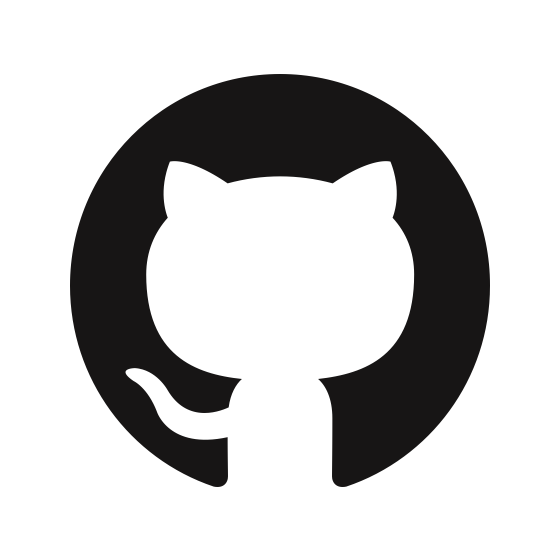}}\hspace{0.1em}Code: \url{https://github.com/Ethylyikes/LatentGeo}.}
\footnotetext[4]{Corresponding author. \texttt{\char60 xuminghu@hkust-gz.edu.cn\char62}.}
\makeatother
\endgroup
\vspace{-0.15cm}

\input{sections/0_abstract}

\input{sections/1_introduction}

\input{sections/2_related_work}

\input{sections/3_method}
\input{sections/4_experiments}

\input{sections/5_conclusion}

\bibliographystyle{splncs04}
\bibliography{main}
\input{sections/6_suppl.tex}

\end{document}

%% file: math_commands.tex
\usepackage{amsmath,amsfonts,bm}

\def\eqref#1{equation~\ref{#1}}
\def\1{\bm{1}}

\DeclareMathAlphabet{\mathsfit}{\encodingdefault}{\sfdefault}{m}{sl}
\SetMathAlphabet{\mathsfit}{bold}{\encodingdefault}{\sfdefault}{bx}{n}

%% file: sections/0_abstract.tex
\begin{abstract}
Despite recent advances in multimodal reasoning, representing auxiliary geometric constructions remains a fundamental challenge for multimodal large language models (MLLMs). Such constructions are absent from the original diagram and must be introduced before theorems apply. Existing approaches predominantly rely on explicit construction paradigms, including text-based geometric specification, visual-token interleaving during reasoning, and tool-augmented geometric execution. However, these methods either fail to faithfully represent complex spatial relationships, incur representation mismatch between discrete symbols and continuous geometric structures, or rely on external capabilities that hinder end-to-end optimization.
To address these limitations, we propose \textbf{LatentGeo}, a framework that learns continuous latent visual representations to internalize auxiliary geometric constructions without pixel-level rendering or external executors. We design a \textbf{three-stage curriculum} that progressively aligns and internalizes these latent representations through auxiliary visual supervision, followed by \textbf{LaGDPO}, a latent-aware reinforcement learning procedure that stabilizes latent representations during policy optimization while improving end-task correctness. To systematically evaluate construction-centric representation quality, we introduce \textbf{\textsc{GeoAux}}, a new benchmark targeting visually dependent geometry problems, and conduct experiments on \textsc{GeoAux} and \textsc{MathVerse}. Results show that LatentGeo achieves substantial gains on geometric reasoning tasks, particularly those requiring auxiliary constructions. Extensive analyses and ablation studies further validate the effectiveness of each component in our framework.

\keywords{Multimodal large language models \and Geometric reasoning \and Latent representations}

\end{abstract}

%摘要中文版

 % 尽管多模态大模型（MLLMs）通过思维链（CoT）在数学推理方面取得了进展，但在需要构建辅助线的几何问题上仍面临严峻挑战。现有的方法主要依赖显式生成，要么调用不稳定的图像编辑工具，要么简单的生成文本描述。然而，前者往往难以精确做出辅助线，后者则因模态差异而丢失了细粒度的空间拓扑信息。虽然隐式推理（Latent Reasoning）在一般视觉任务中已初现潜力，但在复杂的几何规划中尚未得到应用。为了填补这一空白，我们提出了 LatentGeo，这是首个将辅助线构建过程内化为连续隐空间推理的框架。与交替式（interleaved）方法不同，LatentGeo 采用“隐式思维”策略，在推理链之前插入一组可学习的隐式 Token。我们提出了一种新颖的隐式对齐机制（Latent Alignment），利用真实的辅助线图像对这些隐式 Token 进行监督，有效地迫使 model 在生成文本之前在“脑海中”构建解题的“辅助图”。此外，我们引入了三阶段监督微调（SFT）范式并结合强化学习（RL），以稳健地对齐视觉感知与文本推理。在我们的 Geoaux-1600 基准及公开数据集上的广泛实验表明，LatentGeo 在几何推理相关的任务上达到了同尺寸模型的SOTA水平。我们的分析证实，相比于显式生成,隐式推理能够更有效地保留连续的视觉空间语义，从而为解决视觉感知密集型的几何推理问题确立了新的范式。

%% file: sections/1_introduction.tex
\section{Introduction}

\begin{figure*}[!t]
  \centering
\includegraphics[width=1\textwidth]{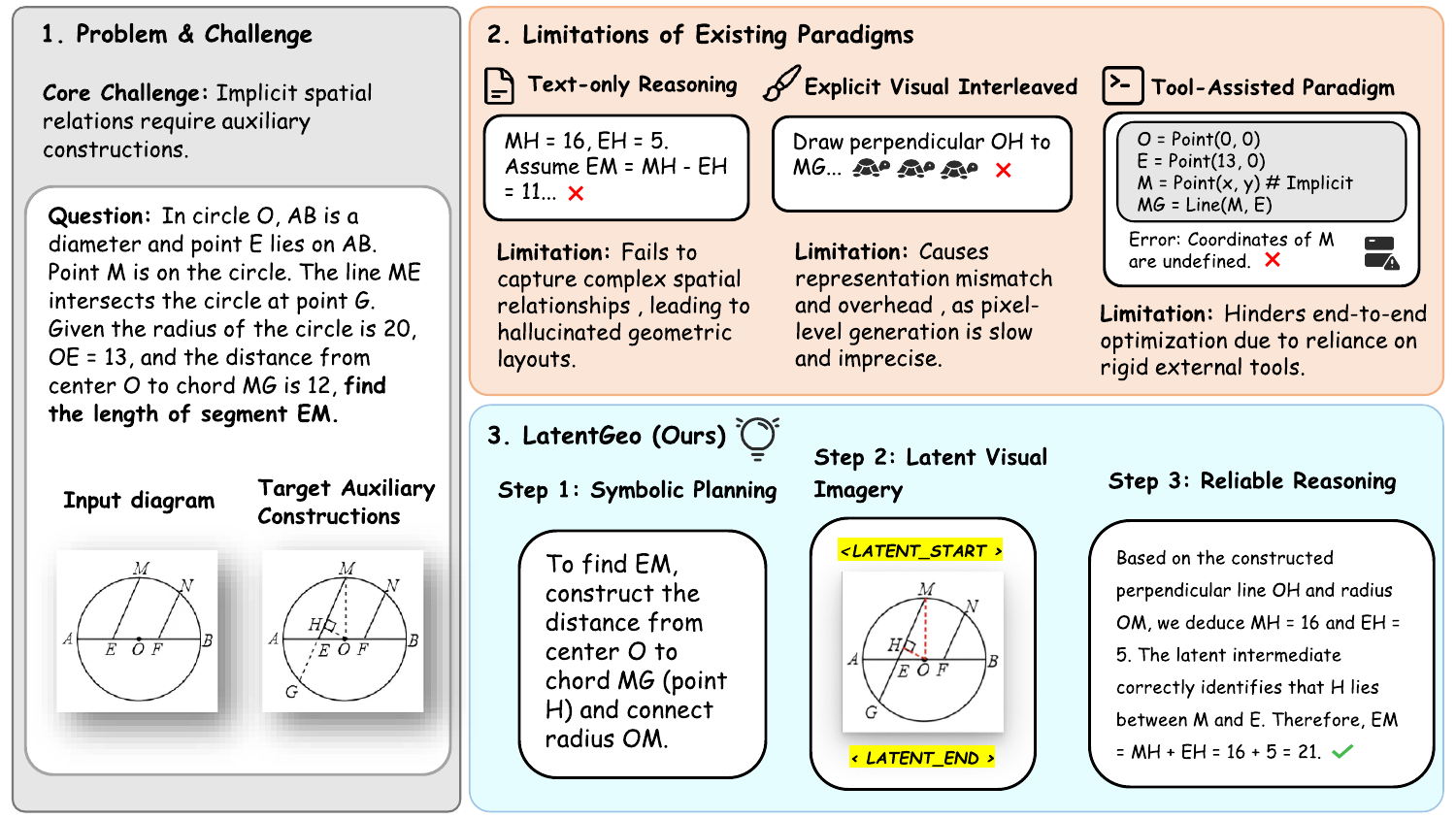}
  \caption{Motivating comparison of auxiliary-construction paradigms for geometric reasoning.}
  \label{fig:case}
\end{figure*}

While multimodal large language models have demonstrated remarkable proficiency in general mathematical reasoning \cite{lu2024mathvista, zhang2024mathverse}, geometry remains a formidable bottleneck. Complex geometric problem-solving frequently relies on auxiliary constructions---new geometric elements absent from the original diagram that must be logically introduced before theorems can be applied. Mastering this process fundamentally tests a model's constructive spatial reasoning and implicit geometric visualization capabilities. However, traditional benchmarks predominantly evaluate the perception of existing visual content, leaving this crucial constructive capability largely under-explored and under-evaluated.

Consequently, existing models exhibit significant shortcomings when tackling such tasks, as illustrated in Fig.~\ref{fig:case}. Current approaches predominantly fall into three paradigms, each suffering from critical limitations. Text-only reasoning often fails to faithfully capture complex spatial relationships, inevitably leading to hallucinated geometric layouts during multi-step deduction. Explicitly interleaving visual operations (e.g., rendering intermediate sketches) \cite{hu2024visualsketchpad} attempts to bridge the modality gap but incurs severe representation mismatch and computational overhead, as pixel-level generation is inherently slow and imprecise. Alternatively, tool-assisted execution paradigms \cite{wang2025mathcodervl} provide precise calculations but hinder end-to-end optimization due to their rigid reliance on external rule-based executors, which frequently fail when geometric constraints are implicit.

To address these fundamental limitations, we propose LatentGeo, a novel reasoning framework that shifts the paradigm from explicit visual generation to latent auxiliary representation. By decomposing the inference process into symbolic planning, latent visual imagery, and reliable reasoning, LatentGeo internalizes auxiliary constructions as a sequence of continuous latent tokens. This approach provides a dedicated, controllable spatial substrate for geometric deduction without the computational burden of pixel-level rendering or the brittleness of external tools.

Equipping models with this latent constructive capability requires a carefully designed training pipeline. We first introduce a three-stage curriculum learning strategy for Supervised Fine-Tuning (SFT). This curriculum progressively aligns text-conditioned latent tokens with genuine geometric structures via auxiliary visual supervision and subsequently internalizes this capability, enabling the model to perform latent constructions autonomously without ground-truth auxiliary diagrams at inference. Furthermore, to directly optimize the model for end-task correctness, we transition to a reinforcement learning (RL) framework. Since standard RL over latent tokens often suffers from severe instability and tends to abruptly degrade into pure textual reasoning, we propose Latent-aware Group-Decoupled Policy Optimization (LaGDPO). LaGDPO explicitly stabilizes the latent representations during policy updates, preventing representation collapse while maximizing the geometric problem-solving reward.

Finally, to systematically diagnose and evaluate the constructive spatial reasoning capabilities of multimodal models, we introduce GeoAux. Unlike existing datasets, GeoAux is a dedicated benchmark specifically targeting complex, visually-grounded geometry problems that necessitate intermediate auxiliary constructions and explicit geometric operations.

Our main contributions can be summarized as follows:
\begin{itemize}
    \item \textbf{Latent Constructive Paradigm:} We propose LatentGeo, a novel framework that internalizes auxiliary geometric constructions as learnable continuous latent tokens, effectively avoiding the overhead of pixel-level rendering and the rigidity of external tools.
    \item \textbf{Curriculum Internalization Strategy:} We design a three-stage SFT curriculum with auxiliary visual supervision to progressively align and internalize latent spatial representations for autonomous diagram-grounded reasoning.
    \item \textbf{Latent-Stabilized Policy Optimization:} We introduce LaGDPO, a tailored reinforcement learning procedure that stabilizes latent visual reasoning and prevents textual degradation while directly optimizing end-task geometric correctness.
    \item \textbf{Construction-Centric Benchmark:} We construct GeoAux, a comprehensive benchmark specifically designed to evaluate the operation-level competence and construction-dependent reasoning capabilities of multimodal models.
\end{itemize}

%% file: sections/2_related_work.tex
\section{Related Work}
% todo: mathematical reasoning
\subsection{Multi-modal Mathematical Reasoning}
% topic
% benchmarks
% methods
Multi-modal mathematical reasoning studies aim to solve mathematical problems that combine textual descriptions with visual contexts such as diagrams and plots.
Representative benchmarks include \textsc{MathVista}~\cite{lu2024mathvista} and \textsc{MathVerse}~\cite{zhang2024mathverse}, which evaluate fine-grained visual perception and diagnose whether models genuinely rely on diagrams, as well as \textsc{MATHVision}~\cite{wang2024mathvision} and \textsc{MV-MATH}~\cite{wang2025mvmath}, which consider competition-style problems and multi-image settings.
For geometry-specific evaluation, \textsc{SOLIDGEO}~\cite{wang2025solidgeo} focuses on solid geometry and 3D spatial reasoning, revealing persistent difficulties in geometric perception, diagram-grounded deduction, and auxiliary constructions.

Motivated by these challenges, recent methods aim to improve explicit multimodal mathematical visual understanding, particularly for geometry-related tasks.
MathCoder-VL~\cite{wang2025mathcodervl} strengthens geometric grounding by aligning diagrams with executable code representations.
E-GPS~\cite{wu2024egps} converts textual and visual inputs into a unified formal representation for goal-driven theorem reasoning.
Geoint-R1~\cite{wei2025geointr1} and GeometryZero~\cite{wang2025geometryzero} incorporate auxiliary constructions through formal verification or learned decision strategies.
Visual Sketchpad~\cite{hu2024visualsketchpad} enables models to externalize intermediate geometric artifacts via explicit sketching actions.
MAVIS~\cite{zhang2025mavis}, selective visual revisitation~\cite{chung2025v1revisitation}, and SVE-Math~\cite{zhang2025openeyes} further enhance token-level grounding and multimodal chain-of-thought reasoning by improving diagram perception and visual evidence reuse during multi-step reasoning.

\subsection{Visual Latent Reasoning}

Latent reasoning performs inference in continuous hidden spaces rather than explicit natural language chains, improving robustness and generalization in complex reasoning tasks~\cite{sun2025lacot_visual}.
This paradigm decouples internal reasoning from surface-level text, mitigating exposure bias and reasoning collapse.

Visual latent reasoning extends latent chain-of-thought to multimodal settings by enabling models to internally manipulate latent visual representations.
This capability is particularly important for visual question answering and diagram-based mathematical reasoning, where spatial and geometric relations are difficult to express in text.

Recent methods introduce latent visual tokens or embeddings as intermediate reasoning states.
Latent-CoT formulates multimodal reasoning as latent variable inference with variational objectives~\cite{sun2025lacot_visual}.
Mirage~\cite{yang2025machinementalimageryempower} and Latent Visual Reasoning~\cite{lvr2025} interleave textual decoding with latent visual embeddings, enabling implicit visual imagination without explicit image generation.

To better support spatial reasoning, Chain-of-Visual-Thought distills dense perceptual cues into compact latent tokens~\cite{qin2025chainofvisualthought}, while Latent Sketchpad enables iterative refinement of latent sketches~\cite{zhang2025latentsketchpad}.
Most recently, Monet unifies multimodal chain-of-thought distillation and reinforcement learning over latent visual tokens, achieving consistent gains on visual perception tasks such as fine-grained real-world scene and document understanding~\cite{wang2025monetreasoninglatentvisual}. 
Despite these advances, existing latent visual reasoning methods uniformly target the \emph{perceptual} challenge of interpreting visual content already present in the input.
The \emph{constructive} challenge unique to geometry---synthesizing auxiliary elements absent from the original diagram before any theorem applies---remains unaddressed, and current evaluation protocols lack dedicated benchmarks to diagnose this capability.

%% file: sections/3_method.tex
%简短的章节引入，介绍method的下面所有部分
\section{Method}
\begin{figure*}[t]
    \centering
    \includegraphics[width=1\linewidth]{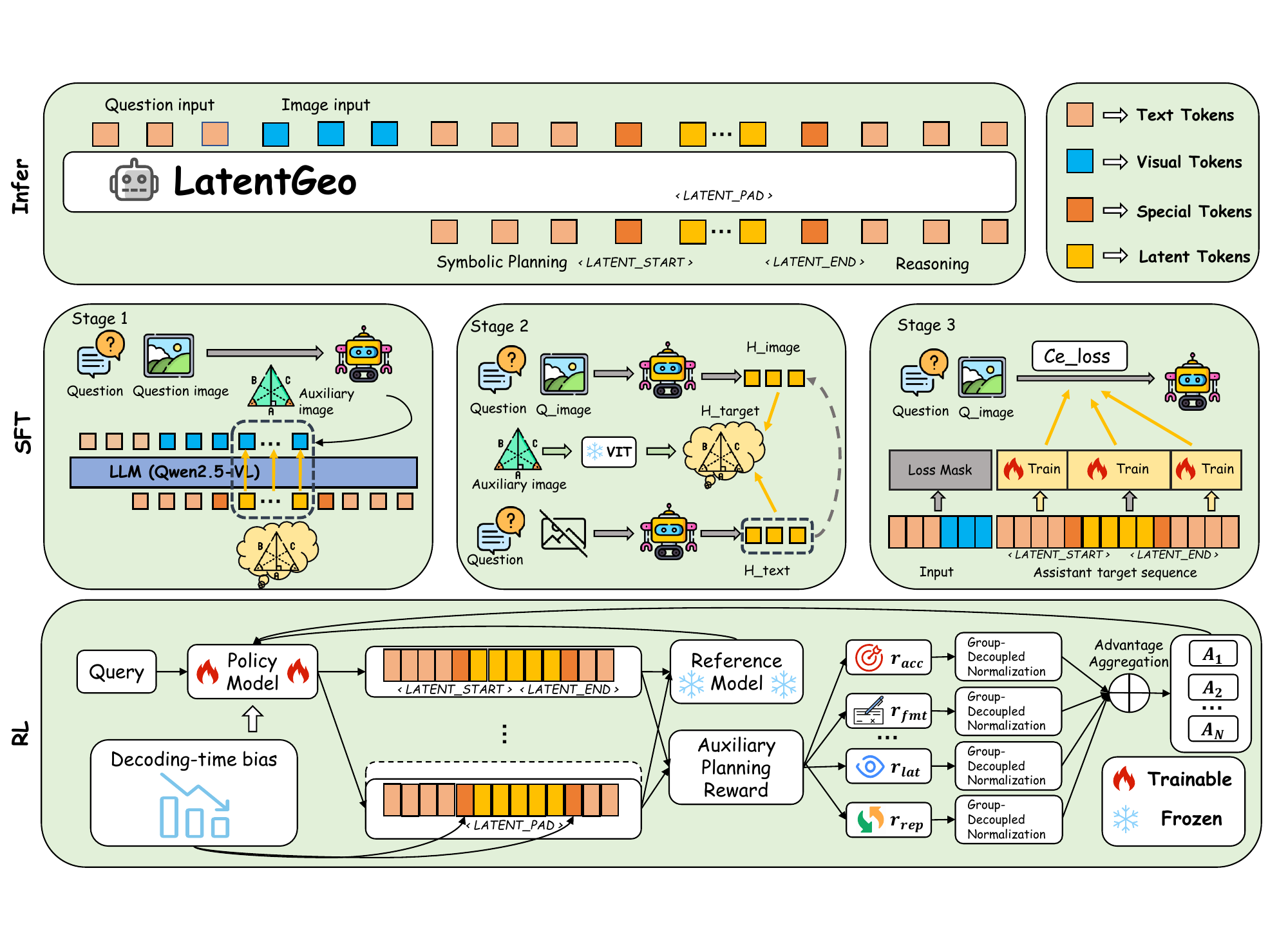}
    \caption{Overview of the \textbf{LatentGeo} framework.}
    \label{fig:framework}
\end{figure*}

% Brief roadmap of the method section.
In this section, we present the full design of \textbf{LatentGeo} for visually grounded geometric reasoning with auxiliary constructions. Unlike perceptual tasks that interpret existing visual content, our focus is on \emph{constructive} geometric operations that introduce new elements into the reasoning process. We begin with the problem formulation and our latent planning factorization, which decomposes inference into \emph{symbolic planning}, \emph{latent construction}, and \emph{final reasoning}, as illustrated in Fig.~\ref{fig:framework}. We then describe the \emph{latent visual thinking} mechanism, where learnable latent tokens are grounded via a projector and trained with a hybrid alignment objective. Subsequently, we introduce a three-stage curriculum learning strategy that progressively internalizes auxiliary construction ability when ground-truth auxiliary diagrams are unavailable at inference time. Finally, we detail \textbf{LaGDPO}, a latent-aware group-decoupled policy optimization procedure that improves end-task utility with KL-regularized updates while preserving the learned latent representations and inference paradigms.

\subsection{Problem Formulation \& Latent Planning Framework}
\label{sec:framework}

Geometric problem solving often requires auxiliary elements that are not explicitly present in the input diagram $\mathcal{I}$. Directly modeling $P(\mathcal{A}\mid \mathcal{I},\mathcal{Q})$ (with question $\mathcal{Q}$ and answer $\mathcal{A}$) is brittle in ``construct-then-reason'' settings, while explicitly generating an auxiliary image $\mathcal{I}_{aux}$ is costly and may introduce visual noise. We instead view auxiliary construction as a plan-driven latent execution process: the model first produces a discrete geometric plan $\mathcal{T}_{plan}$, then generates a continuous latent sequence $\mathcal{Z}$ as an implicit construction intermediate.
Accordingly, we model the inference via the following structured factorization:
\begin{equation}
\label{eq:prob_chain}
P(\mathcal{A}\mid \mathcal{I}, \mathcal{Q}) =
\underbrace{P(\mathcal{T}_{plan}\mid \mathcal{I}, \mathcal{Q})}_{\text{Symbolic Planning}}
\cdot
\underbrace{P(\mathcal{Z}\mid \mathcal{I}, \mathcal{Q}, \mathcal{T}_{plan})}_{\text{Latent Construction}}
\cdot
\underbrace{P(\mathcal{A}\mid \mathcal{I}, \mathcal{Q}, \mathcal{T}_{plan}, \mathcal{Z})}_{\text{Final Reasoning}} .
\end{equation}
 Intuitively, the model first generates $\mathcal{T}_{plan}$ and then autoregressively produces $\mathcal{Z}$; during training, $\mathcal{Z}$ is aligned to visual representations extracted from the ground-truth auxiliary diagram $\mathcal{I}_{aux}$, enabling ``construct-then-reason'' without pixel-level rendering.

\subsection{Latent Visual Thinking Mechanism}
\label{sec:mechanism}

% 这部分对应讲一下论文核心的latent vison方法和原理，对齐的原理，可以介绍一下设计的loss=cos+0.1*mse_loss
To enable the LLM to perform latent auxiliary construction, we modify the architecture of Qwen2.5-VL by introducing a specialized latent thinking mechanism.

\paragraph{Latent Tokens and Projector.}
We expand the model's vocabulary with a set of learnable special tokens
\begin{equation*}
\mathcal{V}_{latent} = \{\, \texttt{<|latent\_start|>},\,\texttt{<|latent\_pad|>},\,\dots,\,\texttt{<|latent\_end|>}\,\}.
\end{equation*}
During the latent visualization phase, the model outputs a sequence of $K$ latent tokens. To ground these tokens in visual reality, we define the \textit{target latent representation} $\mathbf{H}_{target}$ using the ground-truth auxiliary image $\mathcal{I}_{aux}$. 
Specifically, we employ a lightweight projector $\Phi$ to map the visual features of $\mathcal{I}_{aux}$ into the LLM's token embedding space. To maintain spatial semantics while reducing sequence length, we apply a patch-level average pooling strategy:
\begin{equation}
\label{eq:target}
    \mathbf{H}_{target} = \Phi\left(\text{Pool}(f_{enc}(\mathcal{I}_{aux}))\right) \in \mathbb{R}^{K \times D},
\end{equation}
where $f_{enc}$ is the visual encoder (e.g., ViT) and $D$ is the hidden dimension of the LLM.

\paragraph{Alignment Objective.}
Let $\mathbf{H}_{gen} \in \mathbb{R}^{K \times D}$ denote the hidden states generated by the model at the positions of the latent tokens. To ensure $\mathbf{H}_{gen}$ captures the visual structure of the auxiliary lines, we impose a hybrid alignment loss $\mathcal{L}_{align}$ combining Cosine Similarity and Mean Squared Error (MSE):
\begin{equation}
\mathcal{L}_{align} = \lambda_{cos} \left(1 - \frac{\mathbf{H}_{gen} \cdot \mathbf{H}_{target}}{\|\mathbf{H}_{gen}\| \|\mathbf{H}_{target}\|}\right) + \lambda_{mse} \|\mathbf{H}_{gen} - \mathbf{H}_{target}\|^2_2.
\end{equation}
The Cosine term enforces semantic direction alignment, ensuring the ``thought'' represents the correct geometric concept, while the MSE term constrains the magnitude intensity. This approach allows the model to learn a continuous representation of geometry without the burden of decoding pixels.

\subsection{Multi-stage Curriculum Learning Strategy}
\label{sec:training}

Training the probabilistic chain in Eq.~\ref{eq:prob_chain} end-to-end is challenging due to the significant modality gap between text instructions and visual features. We propose a three-stage curriculum learning strategy to progressively equip the model with latent reasoning capabilities.

\paragraph{Stage 1: Visual-Latent Alignment.}
In the initial stage, we establish the core grounding between latent tokens and auxiliary construction semantics.
The LLM backbone, projector $\Phi$, and latent token embeddings are jointly trained, while only the visual encoder (ViT) is kept frozen throughout all stages to preserve pretrained perceptual features.
Given the question image $\mathcal{I}$ and symbolic plan $\mathcal{T}_{plan}$, the model performs a forward pass and produces hidden states $\mathbf{H}_{gen}$ at the latent token positions.
Simultaneously, $\mathcal{I}_{aux}$ is encoded by the frozen ViT and compressed via $\Phi$ into the target representation $\mathbf{H}_{target}$ (Eq.~\ref{eq:target}).
The objective is to minimize $\mathcal{L}_{align}$, driving the LLM's hidden space at latent positions to encode the visual semantics of auxiliary constructions.
Crucially, by optimizing the full LLM rather than a shallow adapter, this stage shapes the model's internal representation space to natively support constructive geometric reasoning.

\paragraph{Stage 2: Plan-Guided Latent Internalization.}
Stage~1 establishes visual grounding under full auxiliary supervision, but $\mathcal{I}_{aux}$ is unavailable at inference. Stage~2 bridges this gap by training the model so that $\mathcal{T}_{plan}$ becomes a reliable semantic controller for latent construction. We run two parallel forward passes per batch, both supervised against $\mathbf{H}_{target}$ computed from $\mathcal{I}_{aux}$ via the frozen ViT and $\Phi$:
\begin{itemize}
    \item \textbf{Visual-Conditioned Stream:} Receives $\mathcal{I}$ and $\mathcal{T}_{plan}$, producing $\mathbf{H}_{img}$ supervised by $\mathcal{L}_{sim,img} = \mathcal{L}_{align}(\mathbf{H}_{img}, \mathbf{H}_{target})$. This stream serves as a well-anchored reference signal.
    \item \textbf{Plan-Conditioned Stream:} Receives only $\mathcal{T}_{plan}$ with no image input, removing reliance on visual features and forcing the plan to be the primary driver of latent construction. It produces $\mathbf{H}_{txt}$ supervised by $\mathcal{L}_{sim,txt} = \mathcal{L}_{align}(\mathbf{H}_{txt}, \mathbf{H}_{target})$.
\end{itemize}
An asymmetric consistency loss further aligns $\mathbf{H}_{txt}$ toward $\mathbf{H}_{img}$, with stop-gradient applied to the latter's latent states to prevent collapse:
\begin{equation}
\mathcal{L}_{cons} = 1 - \cos\!\left(\mathbf{H}_{txt},\; \mathrm{sg}(\mathbf{H}_{img})\right).
\end{equation}
A small CE term stabilizes language capability during representation alignment. The full Stage~2 loss is:
\begin{equation}
\mathcal{L}_2 = 0.1\,\mathcal{L}_{CE} + \mathcal{L}_{sim,img} + \mathcal{L}_{sim,txt} + \mathcal{L}_{cons}.
\end{equation}
By end of Stage~2, the model reduces its dependence on $\mathcal{I}_{aux}$ and improves the plan-conditioned controllability of the latent space.

\paragraph{Stage 3: End-to-End Reasoning.}
In the final stage, all auxiliary supervision is removed. The model is trained end-to-end with pure cross-entropy over the complete assistant output, covering $\mathcal{T}_{plan}$, $\mathcal{Z}$, and $\mathcal{A}$ in a single unified sequence. Latent tokens participate in this CE objective alongside the reasoning text---what is absent is the visual alignment signal $\mathcal{L}_{align}$, not token-level supervision. Training and inference follow an identical generation protocol, completing the transition from guided representation learning to fully autonomous constructive reasoning.

\subsection{LaGDPO: Latent-aware Group Decoupled Policy Optimization}
\label{sec:rl}

While prior supervised training establishes fundamental latent reasoning capabilities, token-level optimization inherently diverges from the ultimate objective of geometric problem solving. To directly align the model with final answer correctness and reliable latent construction, we transition to a reinforcement learning framework through policy optimization with Kullback-Leibler regularization. To achieve these specific training targets while preventing degenerate generation behaviors, we design a tailored auxiliary planning reward balanced by a group decoupled advantage estimation strategy. Finally, to prevent the latent thinking paradigm from degrading into pure textual reasoning during high temperature exploration, we incorporate latent aware decoding stabilization.

\paragraph{Policy Optimization with KL Regularization.}
To directly optimize the model for our ultimate geometric reasoning objectives rather than step-wise likelihood, we employ a reinforcement learning framework. Given an input pair $x=(\mathcal{I},\mathcal{Q})$, the actor $\pi_{\theta}$ generates a complete response $y$ that includes the symbolic plan $\mathcal{T}_{plan}$, a latent segment delimited by \texttt{<|latent\_start|>} and \texttt{<|latent\_end|>}, and the final answer $\mathcal{A}$. We optimize $\pi_{\theta}$ using a PPO-style clipped policy gradient objective with a reference policy $\pi_{\mathrm{ref}}$ initialized from the previous stage checkpoint to tightly constrain policy drift:
\begin{equation}
\label{eq:rl_obj}
\mathcal{L}_{\text{LaGDPO}}(\theta) = \mathcal{L}_{\text{PG}}(\theta) + \beta\,\mathcal{L}_{\text{KL}}(\theta),
\end{equation}
where $\beta$ is a fixed coefficient and $\mathcal{L}_{\text{KL}}$ uses a low variance token-wise KL penalty between $\pi_{\theta}$ and $\pi_{\mathrm{ref}}$. For the policy gradient term, we adopt an asymmetric clipped surrogate with dual clipping for negative advantages. Let $r_t(\theta)=\exp\!\big(\log \pi_{\theta}(y_t|x,y_{<t})-\log \pi_{\theta_{\text{old}}}(y_t|x,y_{<t})\big)$ be the probability ratio at token $t$, and $A_t$ be the advantage. The per-token loss is
\begin{equation}
\label{eq:ppo_clip}
\ell_{\text{PG}}(t) = 
\begin{cases} 
\max\left(-r_t(\theta)A_t, -\mathrm{clip}(r_t(\theta), 1-\epsilon_l, 1+\epsilon_h)A_t\right), & \text{if } A_t \ge 0 \\
\min\left(\max(-r_t(\theta)A_t, -\mathrm{clip}(r_t(\theta), 1-\epsilon_l, 1+\epsilon_h)A_t), -cA_t\right), & \text{if } A_t < 0
\end{cases}
\end{equation}
and $\mathcal{L}_{\text{PG}}$ is obtained by averaging $\ell_{\text{PG}}(t)$ over response tokens.

\paragraph{Auxiliary-Planning Reward.}
To explicitly drive the model toward our dual goals of accurate problem solving and stable latent planning while strictly preventing the generation of repetitive or meaninglessly long texts, we design a comprehensive outcome reward system. Concretely, we compute a scalar reward $R(x,y)$ at the end of the response as
\begin{equation}
\label{eq:reward_sum}
R(x,y)= r_{\text{acc}}(y,\mathcal{A}^*) + r_{\text{fmt}}(y) + r_{\text{lat}}(y) + r_{\text{len}}(y) + r_{\text{rep}}(y),
\end{equation}
where $\mathcal{A}^*$ is the ground-truth answer. $r_{\text{acc}}\in\{0,1\}$ parses the predicted answer from the \texttt{\textbackslash boxed\{\}} region when present or otherwise relies on a language model extraction and checks equivalence against $\mathcal{A}^*$, supporting multiple-choice matching and a $\pm 2\%$ tolerance for numeric answers. $r_{\text{fmt}}(y) = 0.2 \cdot \mathbb{I}[\text{``\texttt{\textbackslash boxed}''} \in y]$ lightly incentivizes a well-formed final answer. For latent planning, let $n_{\text{lat}}(y)$ be the number of well-formed latent segments delimited by the boundary tokens; we enforce a single auxiliary construction via
\begin{equation}
\label{eq:latent_reward}
r_{\text{lat}}(y)=0.5\,\mathbb{I}[n_{\text{lat}}(y)=1]-0.2\,\mathbb{I}[n_{\text{lat}}(y)\neq 1].
\end{equation}
To directly penalize overly long and uninformative responses, we apply a soft length penalty based on the response token length $\ell(y)$:
\begin{equation}
\label{eq:length_reward}
r_{\text{len}}(y)=-\lambda_{\text{len}}\cdot \min\!\left(1,\max\!\left(0,\frac{\ell(y)-0.81\ell_t}{0.1\ell_t}\right)\right),
\end{equation}
where $\ell_t{=}0.81L_{\max}$ is set from the maximum generation length $L_{\max}$ and $\lambda_{\text{len}}{=}0.2$. Finally, we penalize degenerate repetition using both $n$-gram duplication ratios and long token runs. Let $\rho_n(y)$ be the duplication ratio of $n$-grams defined as the fraction of repeated $n$-grams among all $n$-grams, and $m(y)$ be the maximum length of a consecutive token run. We define
\begin{equation}
\label{eq:rep_reward}
\begin{aligned}
r_{\text{rep}}(y) = \max \Big( & -R_{\max},-\lambda_{\text{rep}} \sum_{n\in\{3,4\}} \min\left(1,\max\left(0,\frac{\rho_n(y)-\tau_n}{1-\tau_n}\right)\right), \\
& -\lambda_{\text{rep}} \min\left(1,\max\left(0,\frac{m(y)-m_0+1}{m_0}\right)\right) \Big),
\end{aligned}
\end{equation}
where we use $\lambda_{\text{rep}}{=}1.2$, $(\tau_3,\tau_4){=}(0.18,0.12)$, $m_0{=}6$, and $R_{\max}{=}2$.

\paragraph{Group-Decoupled Advantage Estimation.}
Because our framework relies on multiple distinct reward signals, direct aggregation can cause gradient instability and dominate updates with a single high magnitude reward. To stabilize training and properly balance these diverse objectives, we introduce a group decoupled strategy. We draw $N$ stochastic samples $\{y_i^{(j)}\}_{j=1}^{N}$ for each prompt $x_i$ and compute rewards for each component in Eq.~\ref{eq:reward_sum}. Instead of training a value function, we use a group-relative estimator that normalizes each reward component within the $N$ samples of the same prompt, then aggregates them. For component $k\in\{\text{acc},\text{fmt},\text{lat},\text{len},\text{rep}\}$, we define
\begin{equation}
\label{eq:gdpo_norm}
\tilde{r}_{i,k}^{(j)} = \frac{r_{i,k}^{(j)}-\mu_{i,k}}{\sigma_{i,k}+\epsilon}, \quad \mu_{i,k} = \frac{1}{N}\sum_{j=1}^{N} r_{i,k}^{(j)}, \quad \sigma_{i,k} = \sqrt{\frac{1}{N}\sum_{j=1}^{N}\big(r_{i,k}^{(j)}-\mu_{i,k}\big)^2},
\end{equation}
and combine them as $s_i^{(j)}=\sum_k \tilde{r}_{i,k}^{(j)}$. We further whiten $s_i^{(j)}$ over the minibatch to obtain the final advantage $A_i^{(j)}$, which is broadcast to all response tokens to form $A_t$. This latent-aware group decoupled estimator yields highly stable updates without an explicit critic and ensures every individual reward component contributes on a comparable scale.

\paragraph{Latent-Aware Decoding Stabilization.}
During reinforcement learning rollouts, high sampling temperatures naturally induce diverse exploration but frequently cause the policy to output fewer special latent inference tokens. This underutilization leads to sparse learning signals for auxiliary construction, ultimately causing the latent thinking paradigm to degrade into pure textual reasoning. To resolve this degradation while ensuring that early stage interventions do not bias the final model output, we apply a dynamic and automatically decaying decoding time bias on the two latent boundary tokens. We maintain an exponential moving average of the batch mean latent reward $\bar{r}_{\text{lat},g}$ at step $g$,
\begin{equation}
\label{eq:latent_bias}
m_g = \rho m_{g-1}+(1-\rho)\bar{r}_{\text{lat},g}, \quad b_g = \max\big(b_{\min},\, b_0\exp(-\lambda m_g)\big),
\end{equation}
and add $b_g$ to the logits of \texttt{<|latent\_start|>} and \texttt{<|latent\_end|>} during sampling. As the model successfully learns to consistently produce a single latent block, which is reflected by a higher $\bar{r}_{\text{lat},g}$, the applied bias decays automatically. This adaptive decay firmly prevents the generation of repeated latent segments under late stage low temperature decoding. 
% Moreover, to faithfully preserve the visual alignment property of the latent span, we mask out tokens inside the latent segment from the policy gradient and Kullback Leibler losses by default. Consequently, the reinforcement learning process primarily optimizes the surrounding textual plan and final reasoning while keeping the essential latent representation strictly stable.

%% file: sections/4_experiments.tex
\section{Experiments}
\definecolor{LightBlueRow}{RGB}{238,246,255}

\begin{table*}[!t]
\centering
\small
\setlength{\tabcolsep}{5pt}
\renewcommand{\arraystretch}{1.3}
\caption{\textbf{Main Results on GeoAux Benchmark.} Performance is measured by Accuracy (Acc \%) on 2,228 instances. We contrast our LatentGeo with proprietary models and state-of-the-art open-source baselines. \textit{Knowledge Domains}: Analytic (Anal.), Euclidean (Eucl.), Functional (Func.), and Spatial (Spat.). \textit{Visual Operations}: Analytic Overlay (Analy.), Angular Construction (Angle), Circular Augmentation (Circ.), Connectivity (Conn.), Transformation (Trans.), Alignment (Align.), Projection (Proj.), and Partitioning (Part.).}
\label{tab:geoaux_main}
\resizebox{\textwidth}{!}{
\begin{tabular}{l|c|c|cccc|cccccccc}
\toprule
\multirow{2}{*}{\textbf{Model}} & \multirow{2}{*}{\textbf{Size}}
& \textbf{Overall}
& \multicolumn{4}{c|}{\textbf{Knowledge Categories (Acc \%)}}
& \multicolumn{8}{c}{\textbf{GeoAux Categories (Acc \%)}} \\
\cmidrule(lr){3-3} \cmidrule(lr){4-7} \cmidrule(lr){8-15}

&
& \textbf{ALL}
& \textbf{Anal.} & \textbf{Eucl.} & \textbf{Func.} & \textbf{Spat.}
& \textbf{Analy.} & \textbf{Angle} & \textbf{Circ.} & \textbf{Conn.} & \textbf{Trans.} & \textbf{Align.} & \textbf{Proj.} & \textbf{Part.} \\
\midrule

% -------------------------------------------------------
% Proprietary Models
% -------------------------------------------------------
\rowcolor{LightBlueRow}
\multicolumn{15}{l}{\textit{\textbf{Closed-source MLLMs}}} \\
GPT-4o-mini~\cite{openai2024gpt4omini}      & -- & 12.2 & 14.3 & 13.9 & 12.4 & 6.2  & 5.3  & 8.3  & 7.3  & 13.1 & 11.1 & 13.3 & 5.5  & 10.8 \\
GPT-4o~\cite{openai2024gpt4o}           & -- & 30.3 & 34.3 & 31.4 & 33.2 & 23.4 & 30.9 & 25.6 & 31.8 & 32.3 & 26.1 & 29.6 & 27.4 & 25.5 \\
Qwen3-VL-Flash~\cite{bai2025qwen3vl}    & -- & 33.4 & 43.8 & 34.1 & 34.3 & 25.2 & 27.7 & 29.3 & 30.9 & 32.8 & 30.5 & 36.5 & 21.2 & 33.1 \\
\midrule

% -------------------------------------------------------
% Open-Source Baselines
% -------------------------------------------------------
\rowcolor{LightBlueRow}
\multicolumn{15}{l}{\textit{\textbf{Open-source MLLMs}}} \\
LLaVA-1.6 (Vicuna-7B)~\cite{liu2024llava}         & 7B  & 7.1  & 5.3  & 6.9  & 9.0  & 8.0  & 7.5  & 5.3  & 5.5  & 6.4  & 7.5  & 6.0  & 13.7 & 6.8 \\
MMR1-7B-RL~\cite{leng2025mmr1}                    & 7B  & 29.7 & 27.6 & 33.9 & 31.5 & 18.3 & 28.7 & 28.6 & 23.6 & 29.1 & 23.9 & 28.4 & 23.3 & 19.3 \\
OpenMMReasoner-RL~\cite{zhang2025openmmreasoner} & 7B  & 10.3 & 12.5 & 10.8 & 16.3 & 5.2  & 9.6  & 8.3  & 10.0 & 9.9  & 11.1 & 8.9  & 5.5  & 5.8 \\
R1-Onevision-7B~\cite{yang2025r1onevision}       & 7B  & 13.2 & 13.6 & 14.2 & 17.4 & 8.4  & 8.5  & 15.0 & 10.0 & 13.0 & 15.0 & 12.2 & 8.2  & 11.9 \\
Qwen2.5-VL-7B~\cite{bai2025qwen25vl}    & 7B  & 14.5 & 19.3 & 13.9 & 22.5 & 10.1 & 12.8 & 6.8  & 9.1  & 14.0 & 18.6 & 15.1 & 11.6 & 12.7 \\
Qwen2.5-VL-32B~\cite{bai2025qwen25vl}   & 32B & 33.1 & 35.5 & 34.9 & 37.1 & 25.2 & 26.6 & 30.1 & 29.1 & 34.4 & 30.5 & 31.7 & 29.5 & 27.7 \\
Qwen3-VL-8B~\cite{bai2025qwen3vl}      & 8B  & 32.7 & 38.5 & 33.3 & 41.0 & 24.5 & 27.7 & 25.6 & 30.9 & 32.2 & 32.3 & 32.7 & 26.7 & 26.7 \\
InternVL2-8B~\cite{chen2023internvl} & 8B & 15.9 & 18.1 & 15.2 & 17.4 & 16.1 & 10.6 & 15.8 & 14.6 & 18.1 & 16.8 & 13.7 & 17.1 & 11.0 \\
InternVL3.5-8B~\cite{wang2025internvl35}   & 8B  & 25.1 & 24.9 & 26.9 & 27.0 & 19.4 & 19.2 & 18.8 & 18.2 & 26.1 & 22.1 & 23.6 & 24.0 & 19.9 \\

\midrule

\rowcolor{LightBlueRow}
\multicolumn{15}{l}{\textit{\textbf{Ours}}} \\
\rowcolor{gray!15}
\textbf{LatentGeo} & 7B
& \textbf{34.6} & 30.6 & \textbf{37.3} & 31.5 & \textbf{30.5}
& 27.7 & \textbf{42.1} & 24.6 & \textbf{34.9} & 27.0 & 31.9 & \textbf{37.0} & \textbf{31.5} \\

\textit{Gain vs Baseline} & --
& \textcolor{teal}{+20.1} & \textcolor{teal}{+11.3} & \textcolor{teal}{+23.4} & \textcolor{teal}{+9.0} & \textcolor{teal}{+20.4}
& \textcolor{teal}{+14.9} & \textcolor{teal}{+35.3} & \textcolor{teal}{+15.5} & \textcolor{teal}{+20.9} & \textcolor{teal}{+8.4}  & \textcolor{teal}{+16.8} & \textcolor{teal}{+25.4} & \textcolor{teal}{+18.8} \\
\bottomrule
\end{tabular}
}
\end{table*}
\begin{table*}[!t]
\centering
\small
% \caption{\textbf{Comparison on \textsc{MATHVERSE} (\textit{testmini}).} We report category-wise results (TD, VI, VD, VO) and set ALL as their average. We focus on visually grounded categories (VI, VD, VO) and include the text-dominant category (TD) as a contrastive reference.}
\caption{\textbf{Comparison on \textsc{MATHVERSE} (\textit{testmini}).} We report category-wise results: Text-Dominant (TD), \textit{Vision-Intensive} (VI, requiring complex visual reasoning), \textit{Vision-Dominant} (VD, where visual information prevails), and \textit{Vision-Only} (VO, without text prompts). Results are averaged in the ALL column. We contrast these visually grounded categories (VI, VD, VO) against the TD reference.}
\resizebox{0.7\textwidth}{!}{
\setlength{\tabcolsep}{4.5pt}
\begin{tabular}{l|c|ccccc}
\toprule
\multirow{2}{*}{\textbf{Model}} & \multirow{2}{*}{\textbf{Size}}
& \multicolumn{5}{c}{\textbf{MathVerse (\textit{testmini})}} \\
\cmidrule(lr){3-7}
& & \textbf{ALL} & \textbf{TD} & \textbf{VI} & \textbf{VD} & \textbf{VO} \\
\midrule

\rowcolor{LightBlueRow}
\multicolumn{7}{l}{\textit{\textbf{Baselines}}} \\
Random & -- & 12.4 & 12.4 & 12.4 & 12.4 & 12.4 \\
Human  & -- & 66.9 & 71.2 & 61.4 & 68.3 & 66.7 \\
\midrule

\rowcolor{LightBlueRow}
\multicolumn{7}{l}{\textit{\textbf{Closed-source MLLMs}}} \\
GPT-4o~\cite{openai2024gpt4o} & -- & 50.5 & 59.8 & 48.0 & 46.5 & 47.6 \\
Gemini-1.5-Pro~\cite{team2024gemini15} & -- & 35.5 & 39.8 & 32.0 & 36.8 & 33.3 \\
Qwen-VL-Plus~\cite{bai2023qwen} & -- & 21.4 & 26.0 & 18.5 & 19.1 & 21.8 \\
\midrule

\rowcolor{LightBlueRow}
\multicolumn{7}{l}{\textit{\textbf{Open-source General MLLMs}}} \\
mPLUG-Owl2-7B~\cite{ye2023mplug} & 7B  & 10.0 & 11.6 & 11.1 & 9.4  & 8.0 \\
InternVL2-8B~\cite{chen2023internvl} & 8B & 32.5 & 39.0 & 32.2 & 30.9 & 27.7 \\
InternVL2.5-8B~\cite{chen2024expandingperformanceboundariesopensource} & 8B & 37.8 & 43.0 & 43.0 & 42.2 & 22.8 \\
DeepSeek-VL~\cite{lu2024deepseek} & 7B & 18.4 & 23.0 & 20.2 & 18.4 & 11.8 \\
SPHINX-V2-13B~\cite{lin2023sphinx} & 13B & 17.3 & 20.8 & 16.4 & 15.6 & 16.2 \\
LLaVA-1.5-13B~\cite{liu2024improved} & 13B & 12.9 & 17.1 & 12.6 & 12.7 & 9.0 \\
LLaVA-NeXT-34B~\cite{liu2024llava} & 34B & 33.9 & 49.0 & 35.2 & 28.9 & 22.4 \\
% Qwen2-VL~\cite{wang2024qwen2} & 7B & 31.8 & 37.4 & 31.3 & 30.3 & 28.1 \\
\midrule

\rowcolor{LightBlueRow}
\multicolumn{7}{l}{\textit{\textbf{Open-source Math MLLMs}}} \\
Math-LLaVA-13B~\cite{shi2024math} & 13B & 22.4 & 27.3 & 24.5 & 21.7 & 16.1 \\
Math-PUMA-Qwen2-7B~\cite{zhuang2024math} & 7B & 33.3 & 42.1 & 33.4 & 31.6 & 26.0 \\
Math-PUMA-DeepSeek-Math~\cite{zhuang2024math} & 7B & 30.8 & 43.4 & 33.6 & 31.6 & 14.7 \\
MAVIS-7B~\cite{zhang2025mavis} & 7B & 34.7 & 43.2 & 34.1 & 29.7 & 31.8 \\
InfiMM-Math~\cite{han24infimm} & 7B & 33.3 & 46.7 & 38.1 & 32.4 & 15.8 \\
MultiMath-7B~\cite{peng2024multimath} & 7B & 26.0 & 34.8 & 28.1 & 25.9 & 15.0 \\
\midrule

\rowcolor{gray!15}
\textbf{LatentGeo} & 7B & \textbf{41.4} &
\textbf{51.4} & \textbf{39.3} & \textbf{37.7} &  \textbf{37.3} \\

\textit{Gain vs SOTA (Open-Source Math)} & -- &
\textcolor{teal}{+6.7} & \textcolor{teal}{+4.7} & \textcolor{teal}{+1.2} & \textcolor{teal}{+5.3} & \textcolor{teal}{+5.5} \\
\bottomrule
\end{tabular}
}
\label{tab:results_on_base}
\end{table*}

\subsection{Experimental Settings}
%这个部分每个小段落的篇幅都不多，简要介绍实验的关键信息
\paragraph{Baselines}
We compare against representative multimodal math-reasoning models and geometry-oriented methods, including both open-source and closed-source baselines, under a unified evaluation protocol.
Detailed baseline configurations and implementation details are provided in the supplementary material.

\paragraph{Training Setups}
%训练阶段分别多少步，设置;评估时的设置
% 内容：sft1 2 3阶段分别训练 5 2 3个epoch，经过比较选择latentsize=10。RL阶段使用...配置进行训练n epoch。所有实验都在四个NVIDIA A800 GPU上进行，附录中有进一步的详细参数分析。
The SFT training is divided into three stages, with the training epochs set to 5, 2, and 5. Regarding the latent configuration, we set the sequence length of the latent imagery to $K=10$ based on empirical ablation. The subsequent RL phase is conducted for only one epoch. All experiments utilize 4$\times$NVIDIA A800 GPUs, with detailed hyperparameters provided in the supplementary material.

\paragraph{Evaluation Benchmarks}
%介绍评估用的benchmark. 我们在多个math 几何的benchmark上做了评估，
We evaluate LatentGeo on two complementary domains using two benchmarks: (1) \textit{Auxiliary Construction Quality} on our proposed GeoAux benchmark, designed to assess the accuracy of intermediate construction planning and its alignment with geometric operations; and (2) \textit{Visual-Dependent Analysis} on MathVerse~\cite{zhang2024mathverse}, which probes the model's sensitivity to fine-grained visual details and diagram understanding. Detailed statistics and split information are provided in the supplementary material.

\subsection{Main Results}

\paragraph{Results on GeoAux.}
Table \ref{tab:geoaux_main} demonstrates that LatentGeo establishes a new state-of-the-art on the GeoAux benchmark with an overall accuracy of 34.6\%, surpassing both proprietary models like GPT-4o and larger open-source models like Qwen2.5-VL-32B. Compared to its foundational backbone of Qwen2.5-VL-7B, our method yields a substantial absolute improvement of 20.1\%. This performance leap is particularly pronounced in complex visual operations, where LatentGeo achieves absolute gains of 35.3\% in Angular Construction, 25.4\% in Spatial Projection, and 20.9\% in Connectivity. These results directly validate the efficacy of our latent visual thinking mechanism and the progressive three-stage curriculum. By learning continuous latent representations instead of relying on expensive pixel-level rendering or brittle external tools, the model successfully internalizes auxiliary constructions to faithfully encode complex spatial relationships.

\paragraph{Results on MathVerse.}
Table \ref{tab:results_on_base} further confirms the robustness of LatentGeo on the MathVerse benchmark, where it achieves an overall accuracy of 41.4\% and outperforms the best open-source math-specialized models by a margin of 6.7\%. Notably, the most significant gains emerge in visually dependent and visually operation-heavy categories, yielding improvements of 5.3\% in VD and 5.5\% in VO. Such improvements directly result from our latent planning factorization and the LaGDPO reinforcement learning phase. Decomposing inference into symbolic planning and stabilized latent construction prevents the model from hallucinating geometric layouts through text-only reasoning.

\begin{table*}[h]
\centering
\small
\caption{Ablation Study Results. Impact of different training stages and components on overall performance (ALL) and category-wise accuracy on GeoAux visual operations. The full model demonstrates superior performance across most categories, particularly in complex constructions like Angular Construction and Spatial Projection.}
\label{tab:ablation_simple}
\resizebox{0.8\textwidth}{!}{
\setlength{\tabcolsep}{4.5pt}
\begin{tabular}{l|c|cccccccc}
\toprule
\multirow{2}{*}{\textbf{Model Variant}}
& \textbf{Overall}
& \multicolumn{8}{c}{\textbf{GeoAux Categories (Acc \%)}} \\
\cmidrule(lr){2-2} \cmidrule(lr){3-10}

& \textbf{ALL}
& \textbf{Analy.} & \textbf{Angle} & \textbf{Circ.} & \textbf{Conn.} & \textbf{Trans.} & \textbf{Align.} & \textbf{Proj.} & \textbf{Part.} \\
\midrule

% 1. 架构变体
LatentGeo (TextOnly)         & 26.7 & 28.7 & 27.1 & 30.0 & 25.3 & 22.1 & 24.3 & 30.8 & 24.3 \\
\midrule

% 2. 训练策略变体
w/o Curric. Step 2         & 13.1 & 12.8 & 19.6 & 6.4  & 12.8 & 7.5  & 12.2 & 13.7 & 14.1 \\
w/o Curric. Step 3         & 23.1 & 24.5 & 28.6 & 20.9 & 22.4 & 22.6 & 21.5 & 21.2 & 20.5 \\
w/o RL Phase               & 32.1 & 22.3 & 34.6 & 27.3 & 32.2 & 22.1 & 29.3 & 37.0 & 23.5 \\
\midrule

% 3. RL 方法变体
w/ GRPO~\cite{shao2024deepseekmath}                  & 26.7 & 20.2 & 34.6 & 27.3 & 27.0 & 23.9 & 24.2 & 24.7 & 20.9 \\
w/ GDPO~\cite{liu2026gdpogrouprewarddecouplednormalization}                   & 26.1 & 18.1 & 30.1 & 23.6 & 25.5 & 25.7 & 24.9 & 30.1 & 22.3 \\
\midrule

% 4. 完整模型
\rowcolor{gray!15}
\textbf{LatentGeo (Full)} & \textbf{34.6}
& \textbf{27.7} & \textbf{42.1} & \textbf{24.6} & \textbf{34.9} & \textbf{27.0} & \textbf{31.9} & \textbf{37.0} & \textbf{31.5} \\
\bottomrule
\end{tabular}
}
\end{table*}

\subsection{Ablation Study}
\label{sec:ablation}
We conduct an ablation study to validate our training stages and architectural components, as summarized in Table \ref{tab:ablation_simple}. 

First, introducing latent visual representations provides a distinct spatial substrate that resolves geometric ambiguity. This elevates the full LatentGeo model to 34.6\% overall accuracy, significantly outperforming the 26.7\% achieved by the text-only variant. The advantage is especially prominent in precision-sensitive operations like angular construction and spatial projection.

Second, our progressive curriculum is essential for internalizing these spatial capabilities. Omitting the plan-guided latent internalization, denoted as without Curriculum Step 2, causes a severe performance drop to 13.1\%. This step is critical for aligning text-conditioned latent tokens with visual primitives. Building on this grounded representation, the end-to-end reasoning phase jointly optimizes the symbolic plan and the latent workspace. Removing this third step degrades performance to 23.1\%, highlighting its importance for multi-step logical consistency.

Finally, our latent-aware reinforcement learning phase maximizes end-task utility. While the model achieves 32.1\% before reinforcement learning, LaGDPO improves this to 34.6\%. Crucially, standard policy optimization methods fail in this multimodal constructive setting. Replacing LaGDPO with standard GRPO or GDPO causes accuracy to plummet to 26.7\% and 26.1\% respectively. Without our specific decoding-time bias, the high-temperature rollout setting inherently causes the policy to omit latent segments. This instability triggers a rapid training collapse where the model degenerates to text-only performance levels. LaGDPO effectively prevents this degradation by stabilizing the generation space while directly optimizing geometric correctness.

\subsection{Mechanism Study}

\begin{figure*}[t]
\centering
\includegraphics[width=1\textwidth]{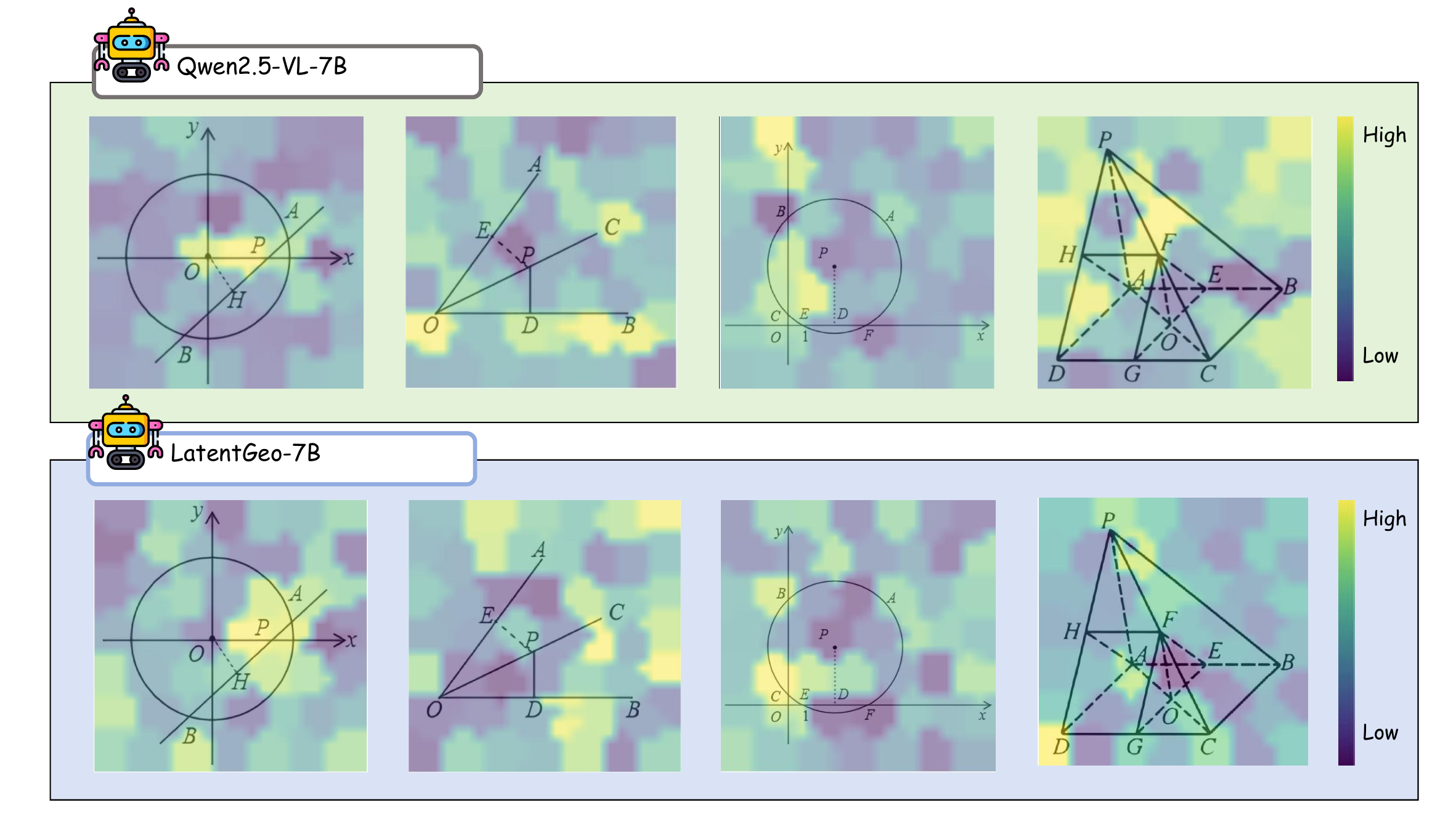}
\caption{Perturbation-based visual attribution on four geometry problems. Submodular insertion and deletion attribution maps are shown for Qwen2.5-VL-7B (top, green) and LatentGeo-7B (bottom, blue).}
\label{fig:attention_map}
\end{figure*}

To further analyze the model's internal mechanism and assess whether the learned representation is effective, we ask: \emph{do latent tokens encode genuine auxiliary construction structure, or merely act as continuation placeholders?}

To probe this, we apply perturbation-based visual attribution~\cite{chen2025whereMllms}: the input image is segmented into superpixels, scored via a greedy submodular insertion and deletion objective, and backprojected to pixels, producing an attribution map of regions most influential to the model's output.

Figure~\ref{fig:attention_map} compares Qwen2.5-VL-7B and LatentGeo-7B across four geometry problems. The baseline exhibits diffuse attributions that frequently include background regions, suggesting limited modeling of the construction-relevant visual operations needed for solving the problems. In contrast, LatentGeo produces structured and localized saliency on regions that act as anchors for auxiliary lines and geometric operations, indicating that our training encourages a learnable latent representation that supports constructive visual reasoning by focusing attention on where auxiliary operations should occur.

Overall, LatentGeo exhibits a construction-aware attribution pattern: its predictions are more consistently grounded in regions that determine feasible constructions rather than incidental background cues. This supports the view that the latent tokens carry meaningful geometric semantics beyond acting as placeholders.

%% file: sections/5_conclusion.tex
\section{Conclusion}

In this paper, we presented LatentGeo, a framework that leverages latent tokens to model auxiliary geometric constructions and improve multimodal geometric reasoning. By shifting from explicit construction paradigms to latent auxiliary representations, LatentGeo mitigates representation mismatch and reduces the reliance on external tools, while remaining amenable to end-to-end optimization. We further introduced a training recipe that combines supervised fine-tuning, reinforcement learning, and auxiliary visual supervision to better align visual perception with textual reasoning. To support systematic evaluation, we introduced GeoAux, a benchmark targeting complex, visually grounded geometry problems requiring intermediate constructions. Extensive experiments on GeoAux and an additional geometry benchmark demonstrate consistent gains, especially on problems where auxiliary constructions are critical. Ablation studies further underscore the necessity of our multi-stage curriculum and LaGDPO for stable latent internalization, while mechanism analysis confirms that the learned latent tokens effectively guide the model's focus toward construction-critical geometric primitives. We hope LatentGeo and GeoAux will facilitate future research on more faithful intermediate representations and more reliable multimodal reasoning for geometry.

% \section{Impact Statement}
% This paper presents work whose goal is to advance the field of Machine Learning by improving multimodal geometric reasoning with intermediate constructions. Our approach may enable more accurate and analyzable geometry problem solving, which could benefit research and educational applications; however, the model may still produce plausible-looking but incorrect constructions that could mislead users if outputs are treated as authoritative, and construction-quality metrics may be susceptible to benchmark-specific optimization. We therefore recommend using the system as an assistive tool with human verification in high-stakes or instructional settings.

%% file: sections/6_suppl.tex
\clearpage
\onecolumn
\raggedbottom
% =========================
% Appendix (Two-Column, A.x)
% =========================
\appendix
\section{Appendix}

% ---------------------------------
% Appendix qualitative-example styles
% ---------------------------------
\definecolor{suppHeader}{HTML}{4F6473}
\definecolor{suppHeaderText}{HTML}{F7F9FB}
\definecolor{suppBody}{HTML}{F5F8FB}
\definecolor{suppBorder}{HTML}{CCD8E2}
\definecolor{suppLabel}{HTML}{354C5E}
\definecolor{suppAccent}{HTML}{5C768A}
\definecolor{suppAccentSoft}{HTML}{DCE6EE}
\definecolor{suppPlaceholder}{HTML}{FBFCFD}

\newtcolorbox{suppexamplebox}[2][]{%
  enhanced,
  colback=suppBody,
  colframe=suppBorder,
  boxrule=0.6pt,
  arc=0.8mm,
  left=2.2mm,
  right=2.2mm,
  top=1.6mm,
  bottom=1.8mm,
  colbacktitle=suppHeader,
  coltitle=suppHeaderText,
  fonttitle=\bfseries\small,
  title={#2},
  toptitle=0.7mm,
  bottomtitle=0.7mm,
  #1
}

\newcommand{\suppfield}[1]{\noindent\textbf{\textcolor{suppLabel}{#1}}}

\newcommand{\suppplaceholder}[2]{%
  \begin{tcolorbox}[
    enhanced,
    colback=suppPlaceholder,
    colframe=suppAccentSoft,
    boxrule=0.5pt,
    arc=0.6mm,
    left=2mm,
    right=2mm,
    top=2mm,
    bottom=2mm,
    width=\linewidth
  ]
  \begin{minipage}[c][#1][c]{0.96\linewidth}
  \centering
  {\small\textcolor{suppAccent}{#2}}
  \end{minipage}
  \end{tcolorbox}
}

% ---------------------------------
% A.1: Detailed statistics / splits
% ---------------------------------
\subsection{Detailed Statistics and Split Information}
\label{app:stats}

We report detailed statistics for the evaluation benchmarks used in this work, including \textsc{MathVerse} and \textsc{GeoAux}.
\paragraph{\textsc{MathVerse}.}
We evaluate on the \textsc{MathVerse} \textit{testmini} split, which comprises four categories: \textbf{TD}, \textbf{VI}, \textbf{VD}, and \textbf{VO}. Each category contains 788 test instances, resulting in a total of 3{,}152 test samples.

\paragraph{\textsc{GeoAux}.} To further assess the model's robustness in \textit{Visually-Dependent Geometric Reasoning}, we curated \textsc{GeoAux}, a diagnostic evaluation suite specifically designed for problems that cannot be solved through direct image-to-text translation. These instances represent a challenging problem scope where the solution remains unreachable through initial visual data alone; instead, the model must introduce external geometric elements---such as auxiliary lines, rotations, or transformations---to bridge the logic gap and complete the reasoning chain.
\begin{table}[h]
\centering
\small
\caption{\textbf{Distribution and taxonomy of the \textsc{GeoAux} benchmark.} Left: taxonomy counts. Right: pie chart. Visual-operation labels are non-exclusive, so their counts may exceed the total number of instances.}
\label{tab:geoaux_stats}
\begin{minipage}[t]{0.45\textwidth}
\vspace{0pt}
\centering
\scriptsize
\renewcommand{\arraystretch}{1.0}
\setlength{\tabcolsep}{2pt}
\begin{tabular}{@{}p{0.76\linewidth}r@{}}
\toprule
\textbf{Statistic} & \textbf{Number} \\
\midrule
\textbf{Total instances} & \textbf{2,228} \\
\midrule
\multicolumn{2}{@{}l@{}}{\textbf{Question type}} \\
-- Free-form & 1,313 \\
-- Multiple-choice & 915 \\
\midrule
\multicolumn{2}{@{}l@{}}{\textbf{Knowledge domain}} \\
-- Analytic coordinate geometry & 265 \\
-- Euclidean plane geometry & 1,320 \\
-- Functional graph geometry & 178 \\
-- Spatial \& projective geometry & 465 \\
\midrule
\multicolumn{2}{@{}l@{}}{\textbf{Visual operation}} \\
-- Analytic overlay & 94 \\
-- Angular construction & 133 \\
-- Circular \& tangential augmentation & 110 \\
-- Elemental connectivity & 957 \\
-- Geometric transformation & 226 \\
-- Orthogonal \& parallel alignment & 1,311 \\
-- Spatial projection \& unfolding & 146 \\
-- Structural partitioning & 498 \\
\bottomrule
\end{tabular}
\end{minipage}\hspace{0.03\textwidth}
\begin{minipage}[t]{0.50\textwidth}
\vspace{0pt}
\centering
\IfFileExists{figures/Geoaux_bing.pdf}{
  \raisebox{-0.05\height}{\includegraphics[width=\linewidth,trim=24bp 11bp 18bp 32bp,clip]{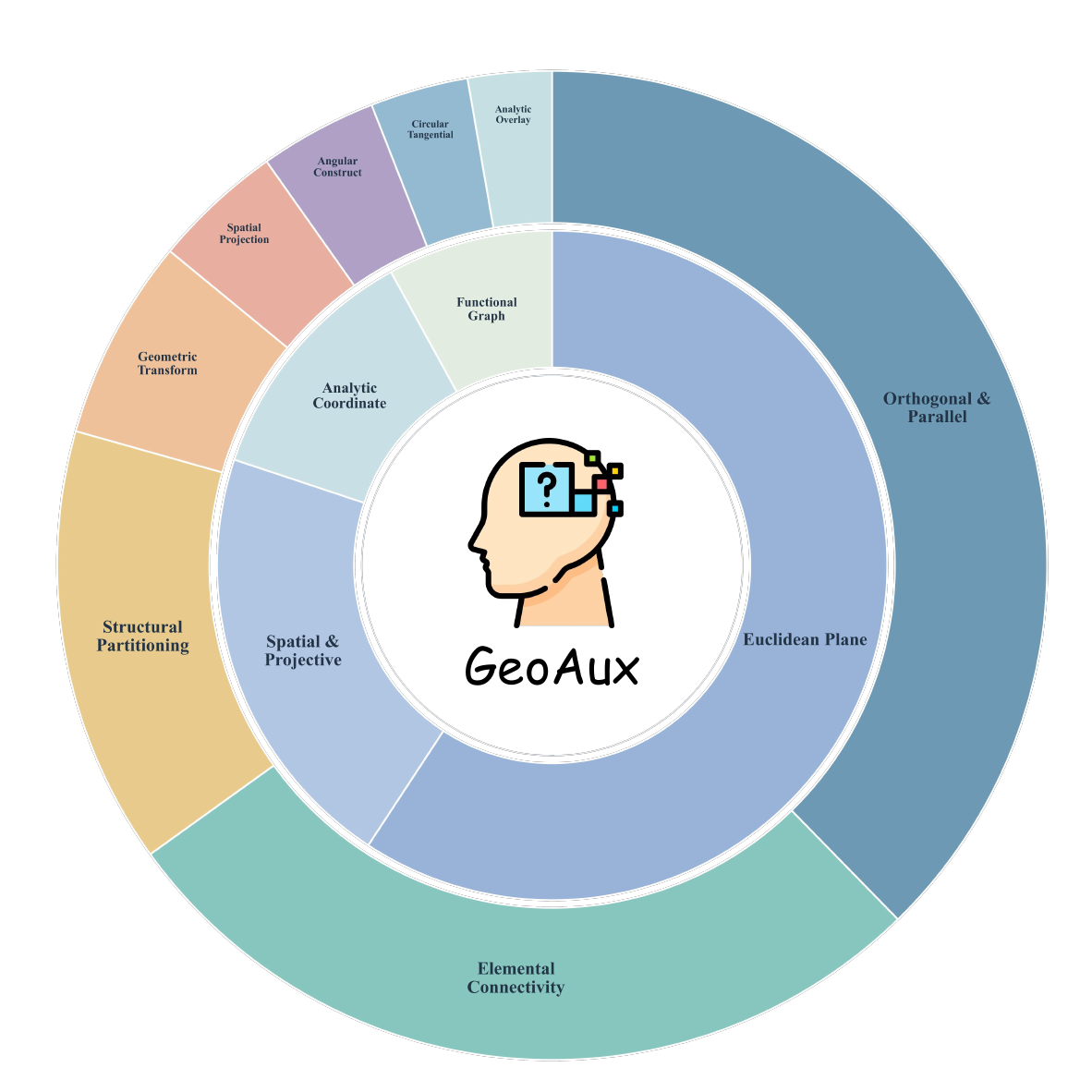}}
}{
  \fbox{\parbox[c][0.62\linewidth][c]{0.92\linewidth}{\centering
  Place your pie chart at\\
  \texttt{figures/Geoaux\_bing.pdf}}}
}
\end{minipage}
\end{table}

\begin{figure}[H]
\centering
\includegraphics[width=0.95\textwidth]{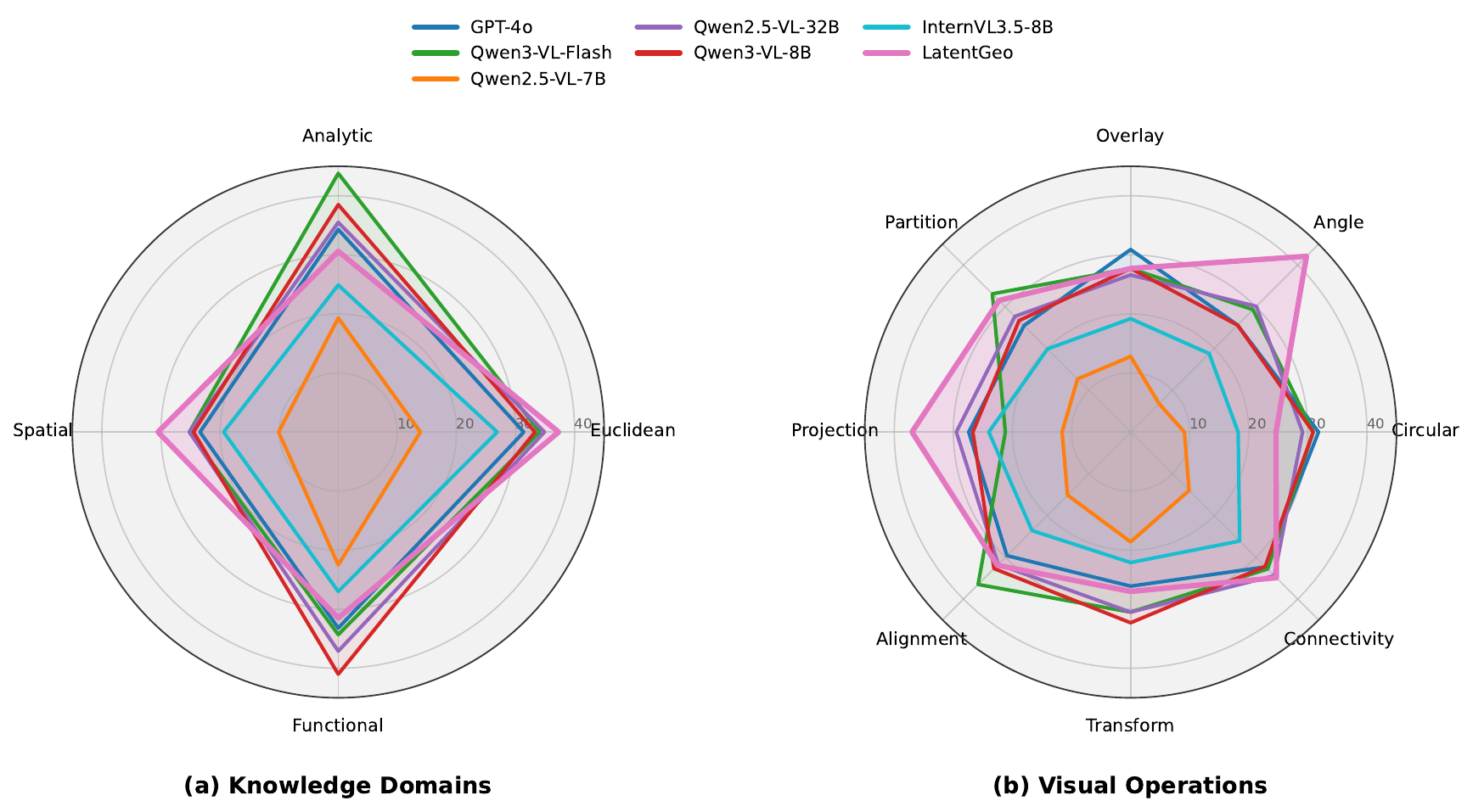}
\caption{\textbf{Radar visualization on \textsc{GeoAux}.} Left: accuracy across four \emph{knowledge domains}. Right: accuracy across eight \emph{visual operations}. Curves are drawn from Table~\ref{tab:geoaux_main} for representative baselines and \textbf{LatentGeo}.}
\label{fig:geoaux_radar}
\end{figure}

The construction of \textsc{GeoAux} begins with an LLM-driven synthesis phase, where frontier models are employed to augment existing geometry problems by proposing complex auxiliary constructions and multi-step constraints. This is followed by a rigorous human-in-the-loop refinement and expert assessment to ensure geometric validity, visual clarity, and to verify that each problem effectively evaluates the intrinsic reasoning capabilities of MLLMs. We categorize each instance based on two dimensions: \textit{Knowledge Domain} and \textit{Visual Operation}. The former represents the mathematical field, while the latter specifies the primary spatial reasoning skill required.

The detailed distribution of the \textsc{GeoAux} benchmark is summarized in Table~\ref{tab:geoaux_stats}. The dataset comprises a total of 2,228 unique instances, with a heavy emphasis on Euclidean Plane Geometry and Alignment-based operations to reflect real-school geometry challenges.

\paragraph{Representative Benchmark Example.}
We present four benchmark exemplars from \textsc{GeoAux}, one for each major knowledge category, to clarify the diversity of auxiliary-construction demands covered by the benchmark.

\begin{figure}[H]
\centering
\begin{suppexamplebox}{Benchmark Example 1}
\small
\suppfield{Knowledge.} Euclidean Plane Geometry.\\
\suppfield{Visual Operation.} Elemental Connectivity.

\vspace{1.5mm}
\suppfield{Question.} Hint: Please solve the problem and provide the final boxed answer as the uppercase option letter only, e.g., \boxed{A}.\\
In the figure, lines $a \parallel b$. In the equilateral triangle $\triangle ABC$, vertex $B$ lies on line $b$, and $\angle 1 = 20^\circ$. Find the measure of $\angle 2$.\\
Options: A. $60^\circ$ \quad B. $45^\circ$ \quad C. $40^\circ$ \quad D. $30^\circ$

\vspace{1.8mm}
\suppfield{Question Image.}
\begin{center}
\IfFileExists{figures/geoaux_3.pdf}{
  \includegraphics[width=0.42\linewidth]{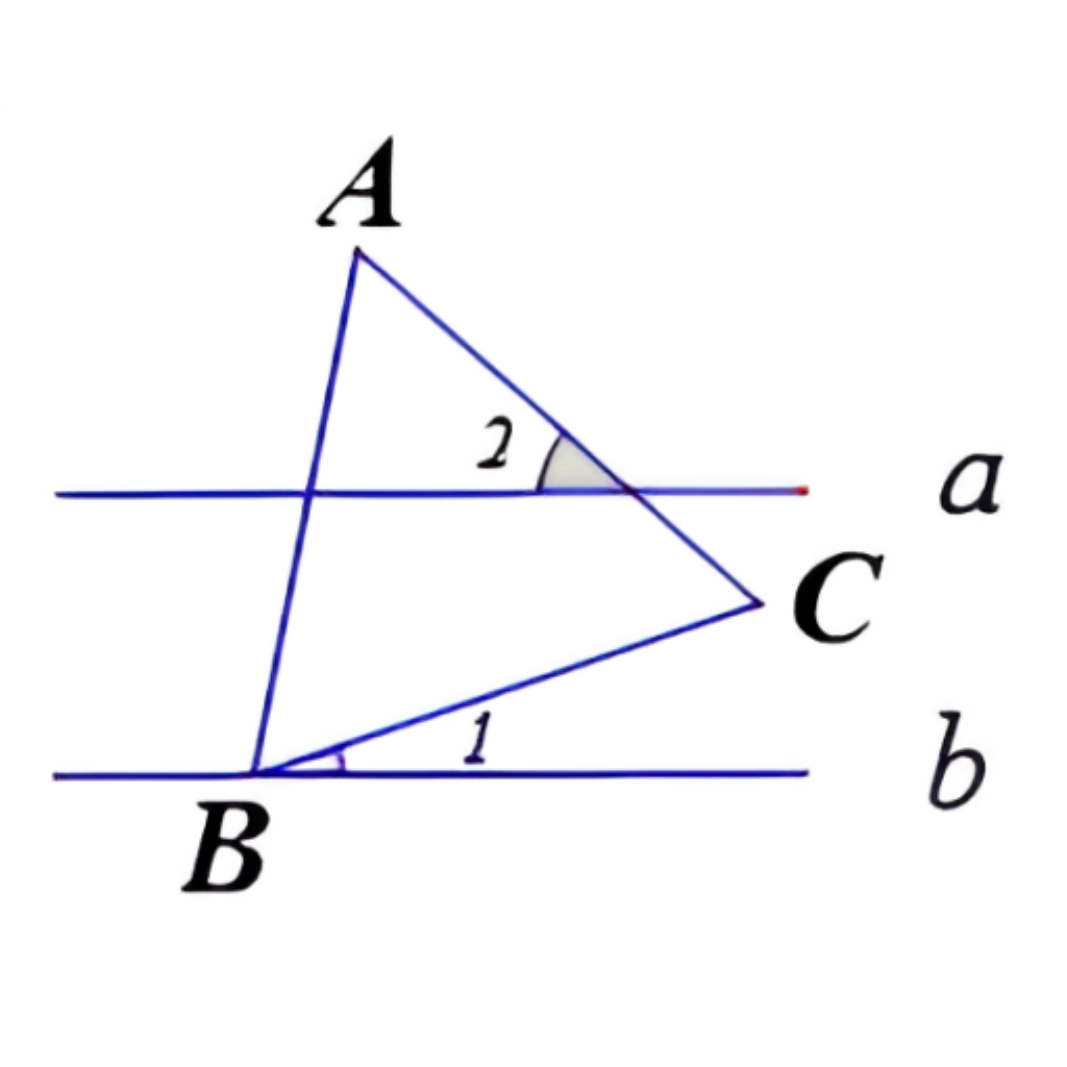}
}{
  \suppplaceholder{0.18\textheight}{Place benchmark image at \texttt{figures/geoaux\_3.pdf}}
}
\end{center}

\vspace{1.2mm}
\suppfield{Construction Requirement.} Extend $AC$ to meet line $b$ and convert the target angle into a solvable relation under the parallel-line constraint.

\vspace{1.2mm}
\suppfield{Correct Answer.} $\boxed{C}$ ($40^\circ$).

\vspace{1.2mm}
\suppfield{Key Evaluation Focus.} This sample tests whether the model can introduce the missing auxiliary connection needed for angle reasoning, rather than relying on direct visual reading alone.
\end{suppexamplebox}
\caption{\textbf{Benchmark example from Euclidean Plane Geometry.}}
\label{fig:benchmark_example_geoaux3}
\end{figure}

\begin{figure}[H]
\centering
\begin{suppexamplebox}{Benchmark Example 2}
\small
\suppfield{Knowledge.} Spatial \& Projective Geometry.\\
\suppfield{Visual Operation.} Orthogonal \& Parallel Alignment; Elemental Connectivity; Structural Partitioning.

\vspace{1.5mm}
\suppfield{Question.} Hint: Please solve the problem and provide the final boxed answer as an exact angle measure, keeping degree notation if needed, e.g., \boxed{50^\circ}.\\
In the cube $ABCD{-}A_{1}B_{1}C_{1}D_{1}$ with edge length $a$, find the angle between the line $A_{1}B$ and the plane $D_{1}B_{1}BD$.

\vspace{1.8mm}
\suppfield{Question Image.}
\begin{center}
\IfFileExists{figures/geoaux_1010.pdf}{
  \includegraphics[width=0.42\linewidth]{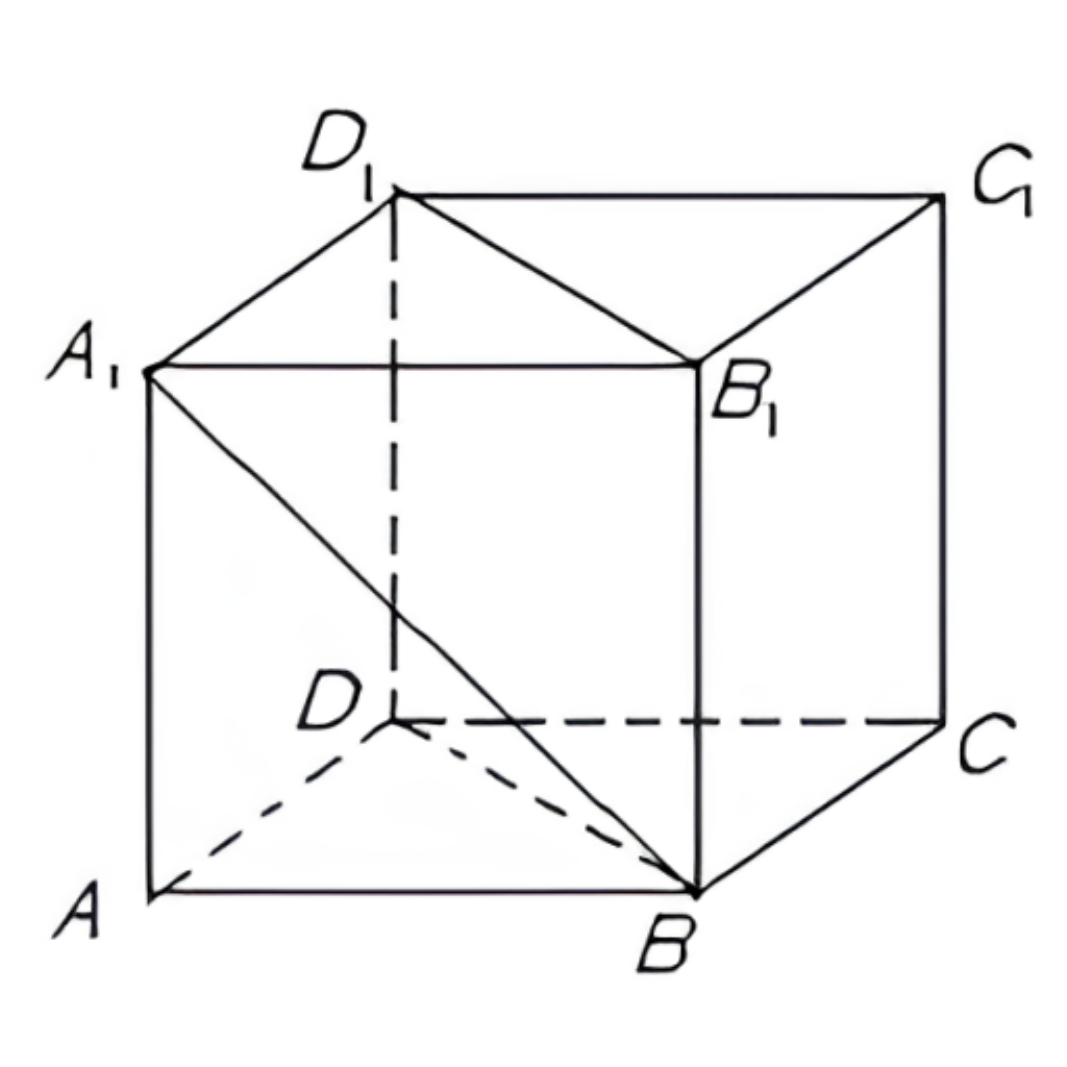}
}{
  \suppplaceholder{0.20\textheight}{Place benchmark image at \texttt{figures/geoaux\_1010.pdf}}
}
\end{center}

\vspace{1.2mm}
\suppfield{Construction Requirement.} Introduce the top-face center via diagonal intersection and identify the foot of the perpendicular from $A_{1}$ to plane $D_{1}B_{1}BD$.

\vspace{1.2mm}
\suppfield{Correct Answer.} $\boxed{30^\circ}$.

\vspace{1.2mm}
\suppfield{Key Evaluation Focus.} This case probes whether the model can decompose a 3D angle-to-plane problem into an auxiliary planar configuration through orthogonality and structural partitioning.
\end{suppexamplebox}
\caption{\textbf{Benchmark example from Spatial \& Projective Geometry.}}
\label{fig:benchmark_example_geoaux1010}
\end{figure}

\begin{figure}[H]
\centering
\begin{suppexamplebox}{Benchmark Example 3}
\small
\suppfield{Knowledge.} Analytic Coordinate Geometry.\\
\suppfield{Visual Operation.} Elemental Connectivity.

\vspace{1.5mm}
\suppfield{Question.} Hint: Please solve the problem and provide the final boxed answer as the uppercase option letter only, e.g., \boxed{A}.\\
In the figure, quadrilateral $OABC$ is a rhombus. Line $CD$ is perpendicular to the $x$-axis with foot $D$. The graph of $y=\dfrac{4}{x}$ passes through $C$ and intersects $AB$ at $E$. Given $OD=2$, what is the area of $\triangle OCE$?\\
Options: A. $2$ \quad B. $4$ \quad C. $2\sqrt{2}$ \quad D. $4\sqrt{2}$

\vspace{1.8mm}
\suppfield{Question Image.}
\begin{center}
\IfFileExists{figures/geoaux_1353.pdf}{
  \includegraphics[width=0.42\linewidth]{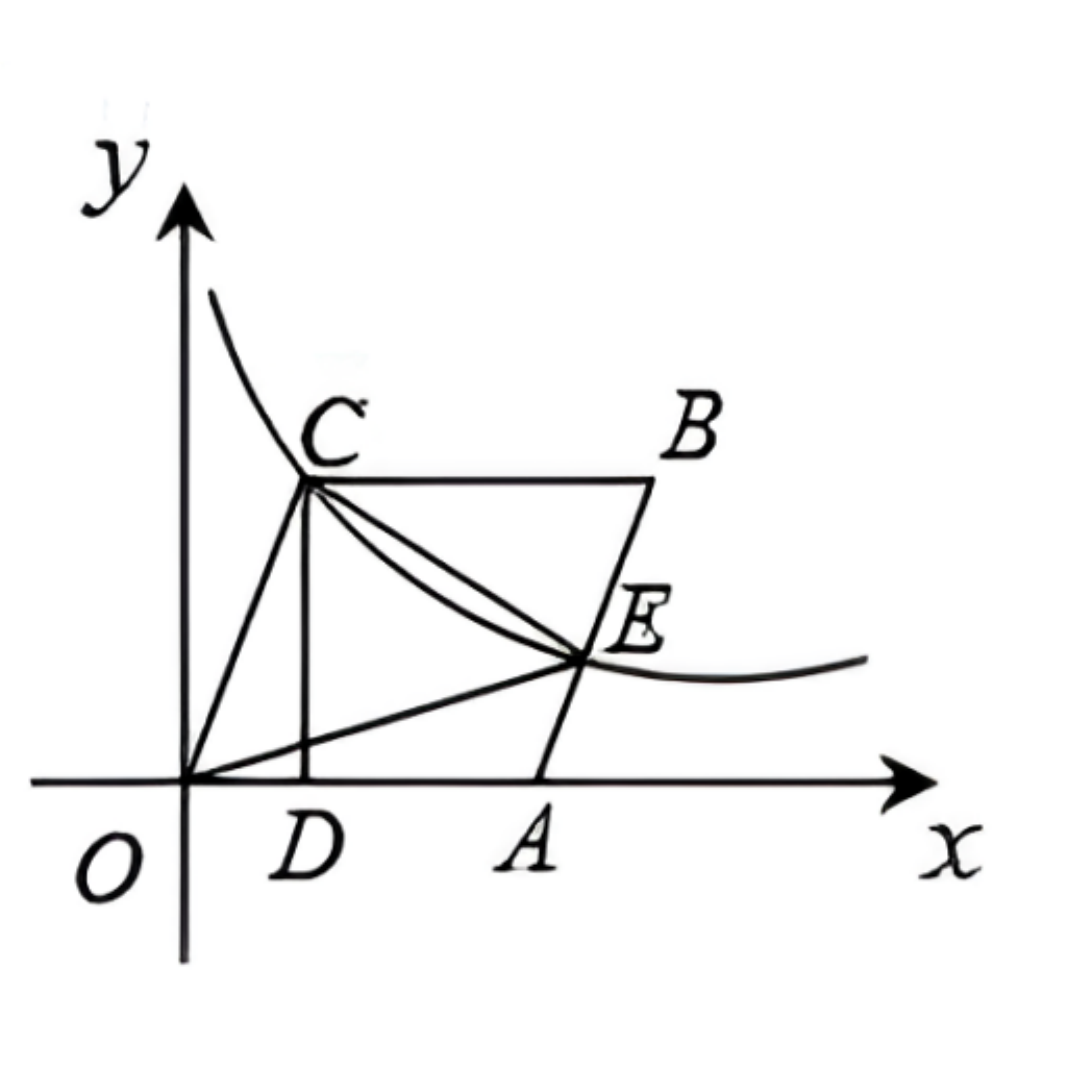}
}{
  \suppplaceholder{0.18\textheight}{Place benchmark image at \texttt{figures/geoaux\_1353.pdf}}
}
\end{center}

\vspace{1.2mm}
\suppfield{Construction Requirement.} Connect $AC$ and combine the conic constraint with the rhombus side-equality relation to recover the geometry needed for the area computation.

\vspace{1.2mm}
\suppfield{Correct Answer.} $\boxed{C}$ ($2\sqrt{2}$).

\vspace{1.2mm}
\suppfield{Key Evaluation Focus.} This sample tests whether the model can bridge symbolic curve information and geometric structure by adding a non-explicit segment that stabilizes the area reasoning.
\end{suppexamplebox}
\caption{\textbf{Benchmark example from Analytic Coordinate Geometry.}}
\label{fig:benchmark_example_geoaux1353}
\end{figure}

\begin{figure}[H]
\centering
\begin{suppexamplebox}{Benchmark Example 4}
\small
\suppfield{Knowledge.} Functional Graph Geometry.\\
\suppfield{Visual Operation.} Analytic Overlay.

\vspace{1.5mm}
\suppfield{Question.} Hint: Please solve the problem and provide the final boxed answer as the uppercase option letter only, e.g., \boxed{A}.\\
As shown, a bridge arch is modeled by a parabola. The maximum height is 16 meters, and the span is 40 meters. On segment $AB$, at a point 5 meters from the center $M$, what is the height of the bridge (in meters)?\\
Options: A. $14$ \quad B. $15$ \quad C. $13$ \quad D. $12$

\vspace{1.8mm}
\suppfield{Question Image.}
\begin{center}
\IfFileExists{figures/geoaux_1633.pdf}{
  \includegraphics[width=0.42\linewidth]{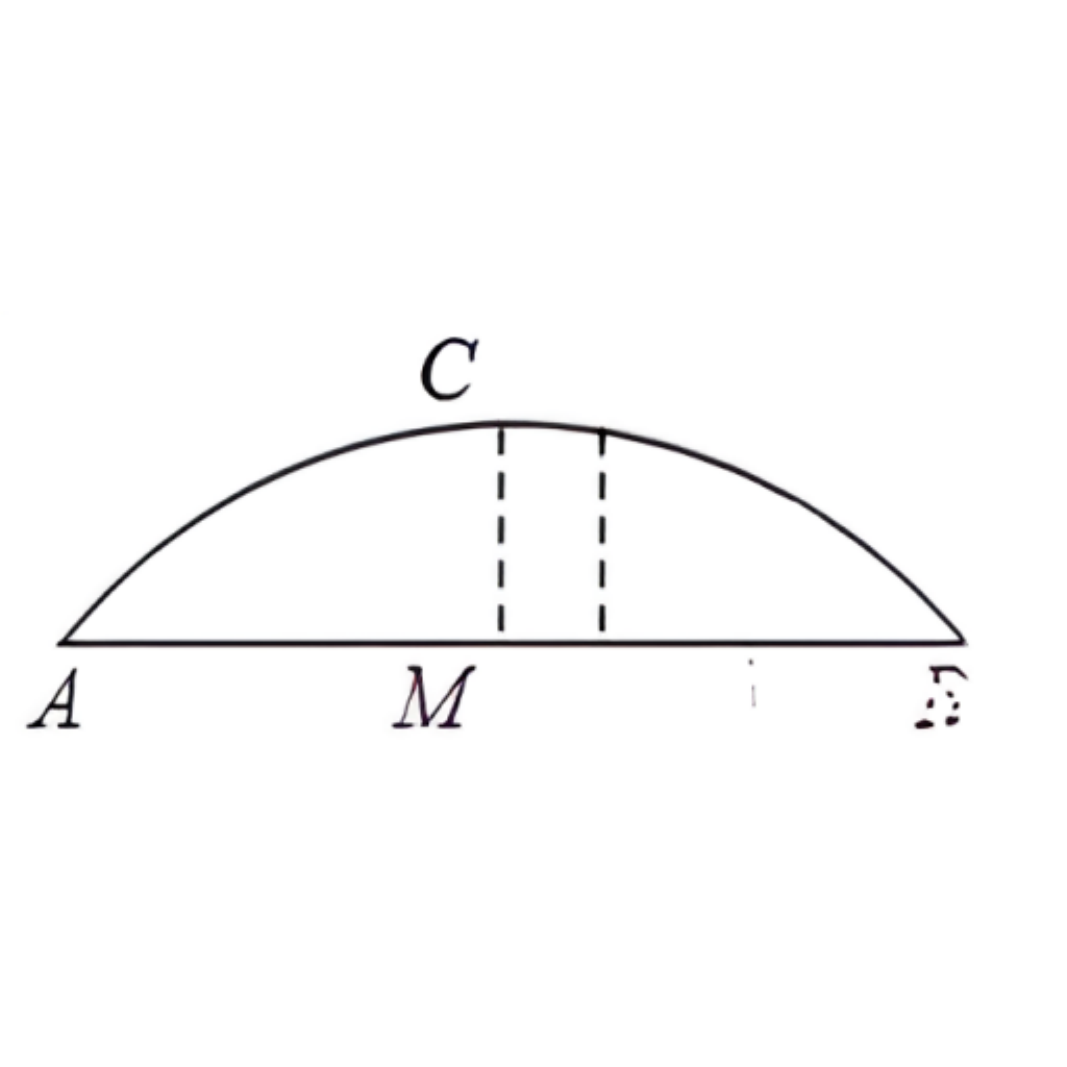}
}{
  \suppplaceholder{0.18\textheight}{Place benchmark image at \texttt{figures/geoaux\_1633.pdf}}
}
\end{center}

\vspace{1.2mm}
\suppfield{Construction Requirement.} Overlay an analytic coordinate system by taking the bridge span as the $x$-axis and the midpoint as the origin, then recover the parabola from its vertex and intercepts.

\vspace{1.2mm}
\suppfield{Correct Answer.} $\boxed{B}$ ($15$).

\vspace{1.2mm}
\suppfield{Key Evaluation Focus.} This sample examines whether the model can translate a real-scene functional graph into an analytic form before performing numeric evaluation.
\end{suppexamplebox}
\caption{\textbf{Benchmark example from Functional Graph Geometry.}}
\label{fig:benchmark_example_geoaux1633}
\end{figure}

% -------------------------------
% A.2: Hyperparameters (SFT + RL)
% -------------------------------
\FloatBarrier
\subsection{Hyperparameters}
\label{app:hparams}

\paragraph{SFT Hyperparameters.}
We fine-tune the model using a multi-stage supervised fine-tuning (SFT) strategy. The learning rate follows a cosine decay schedule, and mixed precision with \texttt{bfloat16} is employed to optimize training efficiency. The detailed hyperparameters for each SFT stage are summarized in Table~\ref{tab:sft_hparams}.

\begin{table}[h]
\centering
\small
\renewcommand{\arraystretch}{1.08}
\setlength{\tabcolsep}{8pt}
\caption{SFT hyperparameters.}
\label{tab:sft_hparams}
\begin{tabular}{@{}p{0.62\textwidth}r@{}}
\toprule
Hyperparameter & Value \\
\midrule
Optimizer & AdamW \\
Learning rate & $1\times 10^{-5}$ \\
Weight decay & 0.01 \\
Max sequence length & 1,024 \\
Latent size & 10 \\
SFT Stage 1 epochs & 5 \\
SFT Stage 2 epochs & 2 \\
SFT Stage 3 epochs & 5 \\
SFT Stage 2 alignment weight & 2.0 \\
SFT Stage 3 alignment weight & 2.0 \\
Training precision & \texttt{bfloat16} \\
\bottomrule
\end{tabular}
\end{table}

\paragraph{RL Hyperparameters.}
We fine-tune the model with LaGDPO, a KL-regularized PPO-style update with group-decomposed advantages.
Table~\ref{tab:rl_hparams} lists the main hyperparameters.
\begin{table}[h]
\centering
\small
\renewcommand{\arraystretch}{1.08}
\setlength{\tabcolsep}{8pt}
\caption{RL hyperparameters.}
\label{tab:rl_hparams}
\begin{tabular}{@{}p{0.62\textwidth}r@{}}
\toprule
Hyperparameter & Value \\
\midrule
KL coefficient $\beta$ & 0.03 \\
PPO epochs & 1 \\
Clip ratio (lower) $\epsilon_l$ & 0.2 \\
Clip ratio (upper) $\epsilon_h$ & 0.3 \\
Dual clip $c$ & 3.0 \\
Optimizer & AdamW \\
Learning rate & $1\times 10^{-7}$ \\
Warmup ratio & 0.15 \\
Max gradient norm & 2.0 \\
Update global batch size & 16 \\
Samples per prompt $N$ & 8 \\
Max prompt length & 4096 \\
Max response length & 2048 \\
Sampling temperature & 0.9 \\
Sampling top-$p$ & 0.99 \\
Length reward max & 0.2 \\
Repetition penalty weight & 1.2 \\
Repetition max penalty & 2 \\
Latent size & 10 \\
Latent logit bias & 10 \\
Latent logit bias decay & 1.0 \\
Latent reward EMA & 0.9 \\
\bottomrule
\end{tabular}
\end{table}

% ---------------------------------
% A.3: Prompting Scheme
% ---------------------------------
\FloatBarrier
\subsection{Prompting Scheme}
\label{app:prompt}

We adopt one unified prompt for SFT/RL/inference and a two-stage evaluation protocol (answer extraction, then answer judging). We report the three prompt templates below.
\begin{figure}[h]
  \centering
  \begin{tcolorbox}[
    title=Prompt: Geometry Reasoning Instruction (SFT \& RL),
    boxrule=0pt,
    left=1mm, right=1mm, top=1mm, bottom=1mm,
    fontupper=\small
  ]
    {\small
    \setlength{\parskip}{4pt}
    \textbf{[Image]}\\
    You are a helpful assistant.\\
    First, analyze the problem with the image and describe all necessary geometric constructions or spatial operations using concise and standard language.\\
    Then, provide a reasoning to solve the problem.\\
    Finally, give the final answer. The final answer must be boxed, e.g., \textbackslash boxed\{answer\}.\\[1mm]
    \textbf{Question:}\\
    \{question\}
    }
  \end{tcolorbox}
  \caption{Unified instruction prompt used for SFT, RL, and inference.}
  \label{fig:geom_prompt}
\end{figure}

\begin{figure}[h]
  \centering
  \begin{tcolorbox}[
    title=Prompt: Answer Extraction for Evaluation,
    boxrule=0pt,
    left=1mm, right=1mm, top=1mm, bottom=1mm,
    fontupper=\small
  ]
    {\small
    \setlength{\parskip}{4pt}
    \textbf{\{DEMO\_PROMPT\}}\\
    (Contains extraction guidelines and in-context examples.)\\[1mm]
    \textbf{\{hint\}}\\
    Hint: This is a Multiple Choice Question. Extract only the option letter (A, B, C, D); if the model outputs option content, map it back to the corresponding letter when possible.\\
    Hint: This is a Free-form Question. Extract the final value or expression.\\[1mm]
    \textbf{Question:} \{query\}\\
    \textbf{Model response:} \{response\}\\
    \textbf{Extracted answer:}
    }
  \end{tcolorbox}
  \caption{Answer-extraction template used before final matching.}
  \label{fig:extract_prompt}
\end{figure}
\begin{figure}[h]
  \centering
  \begin{tcolorbox}[
    title=Prompt: Equivalence-Based Answer Judging,
    boxrule=0pt,
    left=1mm, right=1mm, top=1mm, bottom=1mm,
    fontupper=\small
  ]
    {\small
    \setlength{\parskip}{4pt}
    You are a math teacher grading a student's answer.\\
    Question: \{question\}\\
    Correct Answer: \{ground\_truth\}\\
    Student Answer: \{prediction\}\\[1mm]
    Does the student answer match the correct answer?\\
    For numerical answers, they should be equivalent (e.g., 1.5 equals 3/2).\\
    For multiple choice, the option letter should match.\\
    For expressions, they should be algebraically equivalent.\\[1mm]
    Reply only with ``True'' or ``False''.
    }
  \end{tcolorbox}
  \caption{Judge prompt used for final correctness decision.}
  \label{fig:judge_prompt}
\end{figure}

% --------------------------------
% A.4: Training Data Sources
% --------------------------------
\FloatBarrier
% Supplement-only citations begin in this section.
% These references are not cited in the verified main-text sections.
\subsection{Training Data Sources}
\label{app:training-data}

\paragraph{SFT Training Data.}
We construct a high-quality Supervised Fine-Tuning (SFT) dataset comprising approximately 30,000 geometry problems. These samples are aggregated from multiple public geometric reasoning benchmarks, including \textsc{MathCanvas}, \textsc{Geometry3K}, \textsc{GeoQA+}, \textsc{GEOS}, and \textsc{UniGeo}.
To ensure the data quality and alignment with our training objective, we employ a rigorous data processing pipeline involving both manual curation and LLM-assisted reformatting.
Specifically, raw data instances are first manually filtered to exclude low-quality images or ambiguous problem statements. Subsequently, we utilize advanced LLMs (e.g., GPT-4o) to canonicalize the solution traces into the unified format required for SFT (as detailed in Appendix~\ref{app:prompt}), ensuring a consistent structure of visual description, reasoning steps, and the final boxed answer.

\begin{figure}[H]
\centering
\begin{suppexamplebox}{Representative SFT Training Example}
\footnotesize
\suppfield{Question.} In square $ABCD$ with side length $2$, the angle bisector of $\angle DAC$ intersects $DC$ at $E$. Points $P$ and $Q$ move on $AD$ and $AE$, respectively. Find the minimum value of $DQ+PQ$.

\vspace{1.0mm}
\suppfield{Question Image.}\par
{\centering
\IfFileExists{figures/ques_img.pdf}{
  \includegraphics[width=0.32\linewidth]{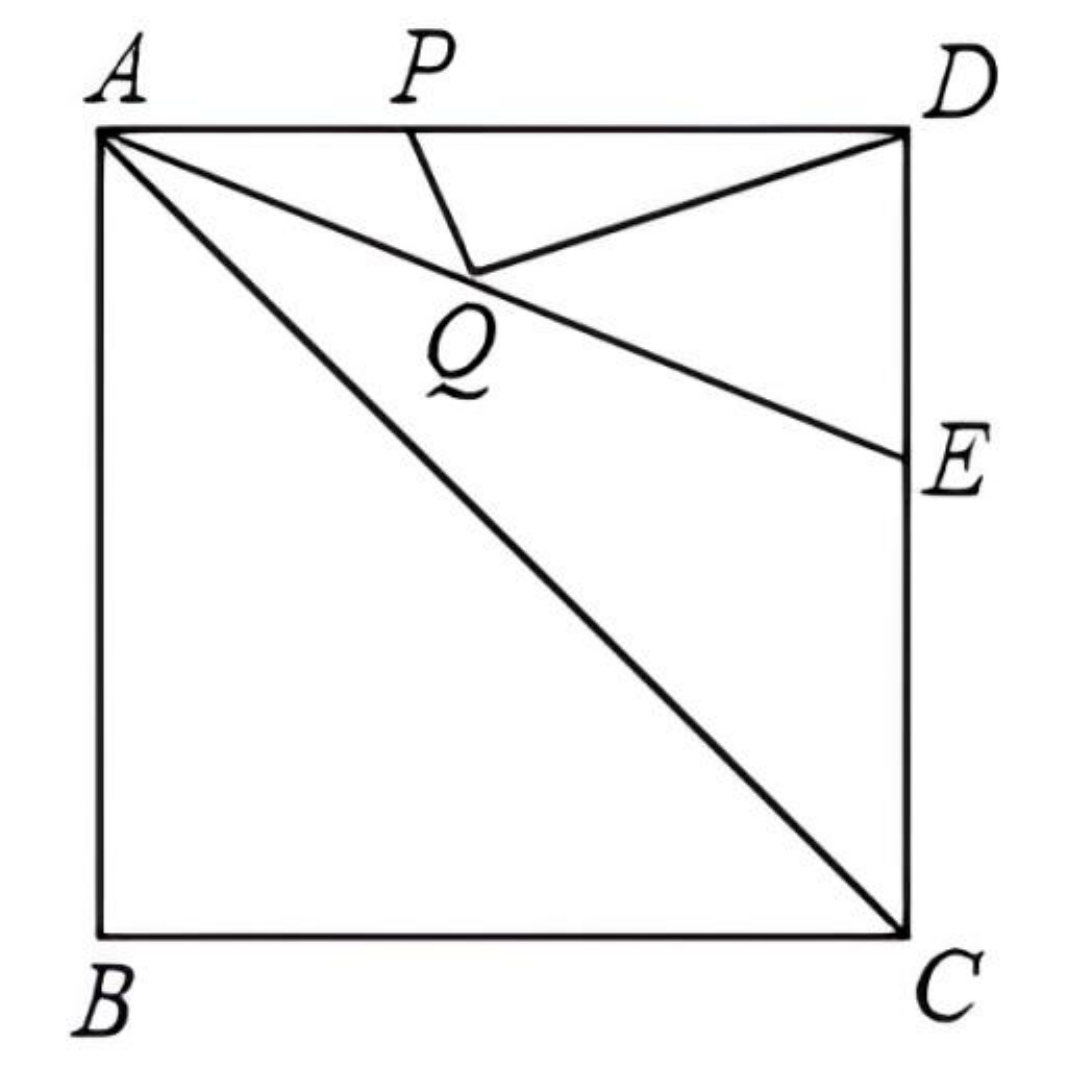}
}{
  \suppplaceholder{0.16\textheight}{}
}
\par}

\vspace{0.8mm}
\suppfield{Assistant Output.} Reflect $D$ across line $AE$ to point $D'$. Since $Q$ lies on $AE$, convert $DQ+PQ$ into the broken path $D'Q+QP$, then reduce the minimization to the distance from $D'$ to line $AD$. Let $P'$ be the foot of the perpendicular from $D'$ to $AD$.

\vspace{3mm}
\par\noindent\texttt{<|latent\_start|><|latent\_pad|>\dots<|latent\_pad|><|latent\_end|>}\par
\vspace{3mm}

Because $Q$ lies on the mirror line $AE$, reflection gives $DQ=D'Q$. Hence $DQ+PQ=D'Q+QP$. For any point $P\in AD$, the broken path from $D'$ to $P$ through $Q$ is at least the straight-line distance $D'P$, with equality when $D',Q,P$ are collinear. Therefore, the problem reduces to minimizing $D'P$ for $P$ on line segment $AD$.\\
This minimum is attained when $P=P'$, the foot of the perpendicular from $D'$ to $AD$, so $\min(DQ+PQ)=D'P'$. Since $AE$ bisects $\angle DAC$, reflecting ray $AD$ across $AE$ maps it to ray $AC$. Therefore, $D'$ lies on $AC$ and $AD'=AD=2$. In a square, diagonal $AC$ makes a $45^\circ$ angle with side $AD$, so $\angle D'AP'=45^\circ$. Thus $\triangle AD'P'$ is a $45^\circ$-$45^\circ$-$90^\circ$ right triangle with hypotenuse $AD'=2$, giving $D'P'=\dfrac{2}{\sqrt{2}}=\sqrt{2}$. Therefore, the minimum value is $\boxed{\sqrt{2}}$.

\vspace{0.8mm}
\suppfield{Auxiliary Image.}\par
{\centering
\IfFileExists{figures/solution_img.pdf}{
  \includegraphics[width=0.30\linewidth]{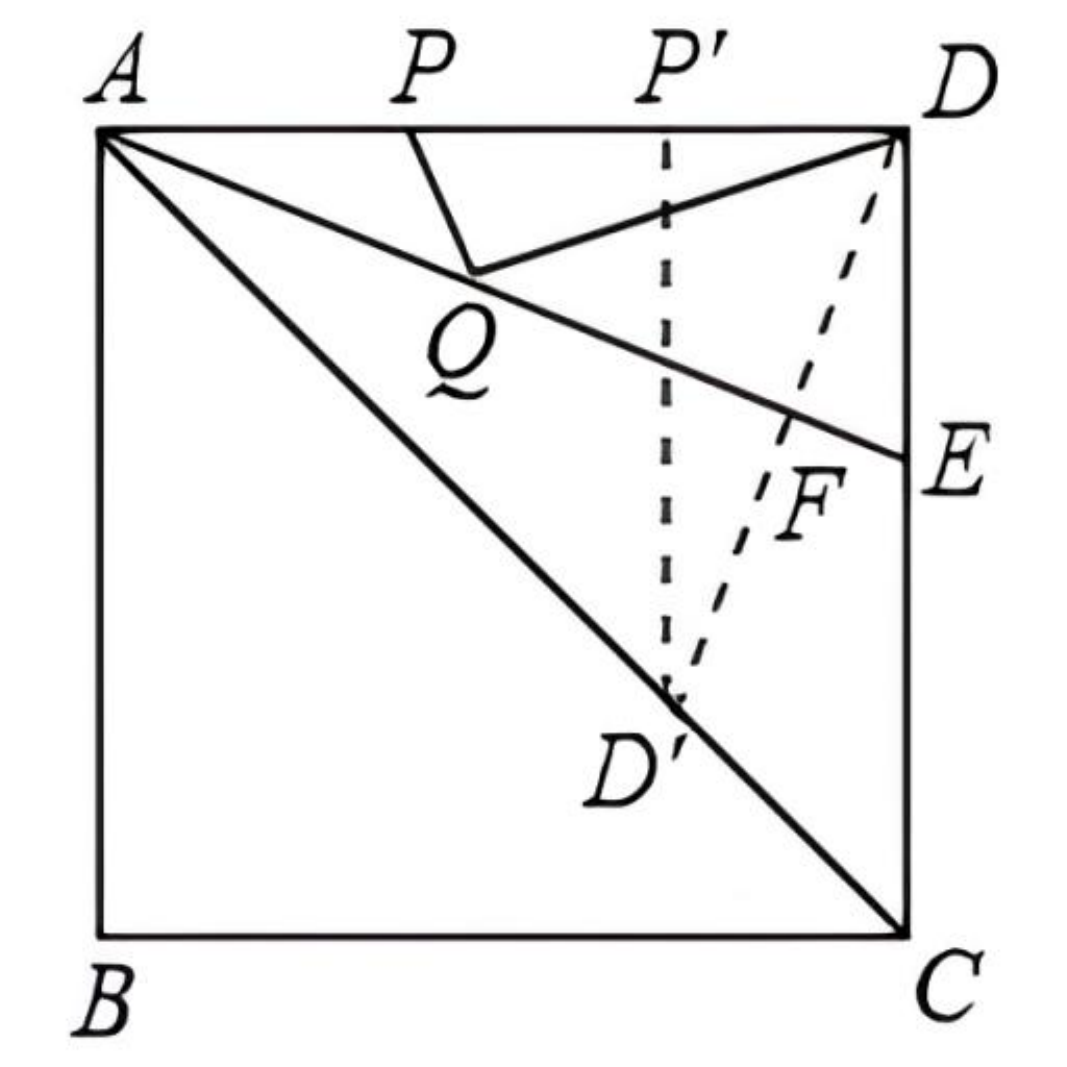}
}{
  \suppplaceholder{0.07\textheight}{}
}
\par}
\end{suppexamplebox}
\caption{\textbf{Representative SFT training example.}}
\label{fig:training_example_template}
\end{figure}

\paragraph{Data Composition.}
The SFT dataset integrates diverse geometric challenges from the following sources:

\begin{itemize}[leftmargin=*, itemsep=0pt]
    \item \textbf{\textsc{MathCanvas}}~\cite{shi2025mathcanvasintrinsicvisualchainofthought}: A multimodal dataset emphasizing intrinsic visual reasoning. We select instances that require detailed visual understanding to complement text-based logic.
    \item \textbf{\textsc{Geometry3K}}~\cite{lu2021intergps}: A comprehensive benchmark of high-school geometry problems. We utilize its rich annotations to construct step-by-step reasoning chains.
    \item \textbf{\textsc{GeoQA+}}~\cite{cao2022mgeo}: A large-scale geometric question-answering dataset. We filter for problems that involve complex calculation and theorem application.
    \item \textbf{\textsc{GEOS}}~\cite{seo2015solving}: One of the pioneering datasets for geometry problem solving, providing fundamental plane geometry instances.
    \item \textbf{\textsc{UniGeo}}~\cite{chen2022unigeo}: A unified geometric reasoning dataset covering both calculation and proof problems across various difficulty levels.
\end{itemize}

\paragraph{RL Training Data.}
Our RL training data are sampled from three public geometry problem-solving datasets: Geometry3K, GeomVerse, and MathCanvas~\cite{lu2021intergps,kazemi2023geomverse,shi2025mathcanvasintrinsicvisualchainofthought}.
In total, we collect 3{,}339 problems, including 1{,}362 from Geometry3K, 1{,}309 from GeomVerse, and 668 from MathCanvas.
For RL, we use each problem (with its associated diagram/image when provided) as the prompt, and treat the dataset-provided final answer as outcome supervision for reward computation.
Importantly, we do not consume intermediate solution traces (e.g., chain-of-thought or step annotations) as supervision in the RL stage.

\paragraph{Data Composition.}
The RL dataset draws from the following sources:

\begin{itemize}[leftmargin=*, itemsep=0pt]
    \item \textbf{\textsc{Geometry3K}}~\cite{lu2021intergps}: A benchmark of 3{,}002 high-school geometry multiple-choice problems. We randomly sample 1{,}362 instances and use the ground-truth answer option as the outcome-based reward signal.
    \item \textbf{\textsc{GeomVerse}}~\cite{kazemi2023geomverse}: A procedurally generated geometry dataset designed to probe reasoning capabilities under controllable factors such as reasoning depth. We sample 1{,}309 problems and use only their final answers for outcome supervision.
    \item \textbf{\textsc{MathCanvas}}~\cite{shi2025mathcanvasintrinsicvisualchainofthought}: A multimodal mathematical reasoning resource emphasizing intrinsic visual reasoning and diagram-centric problem solving. We sample 668 geometry instances and use only the dataset-provided final answers as RL reward supervision.
\end{itemize}

% ------------------------
% A.5: RL Training Curves
% ------------------------
\FloatBarrier
\subsection{RL Training}
\label{app:rl}

Hyperparameters are provided in Appendix~\ref{app:hparams}. We include RL training trend figures in this section to visualize reward evolution during policy optimization.

\begin{figure}[H]
  \centering
  \begin{subfigure}[t]{0.48\textwidth}
    \centering
    \includegraphics[width=\linewidth]{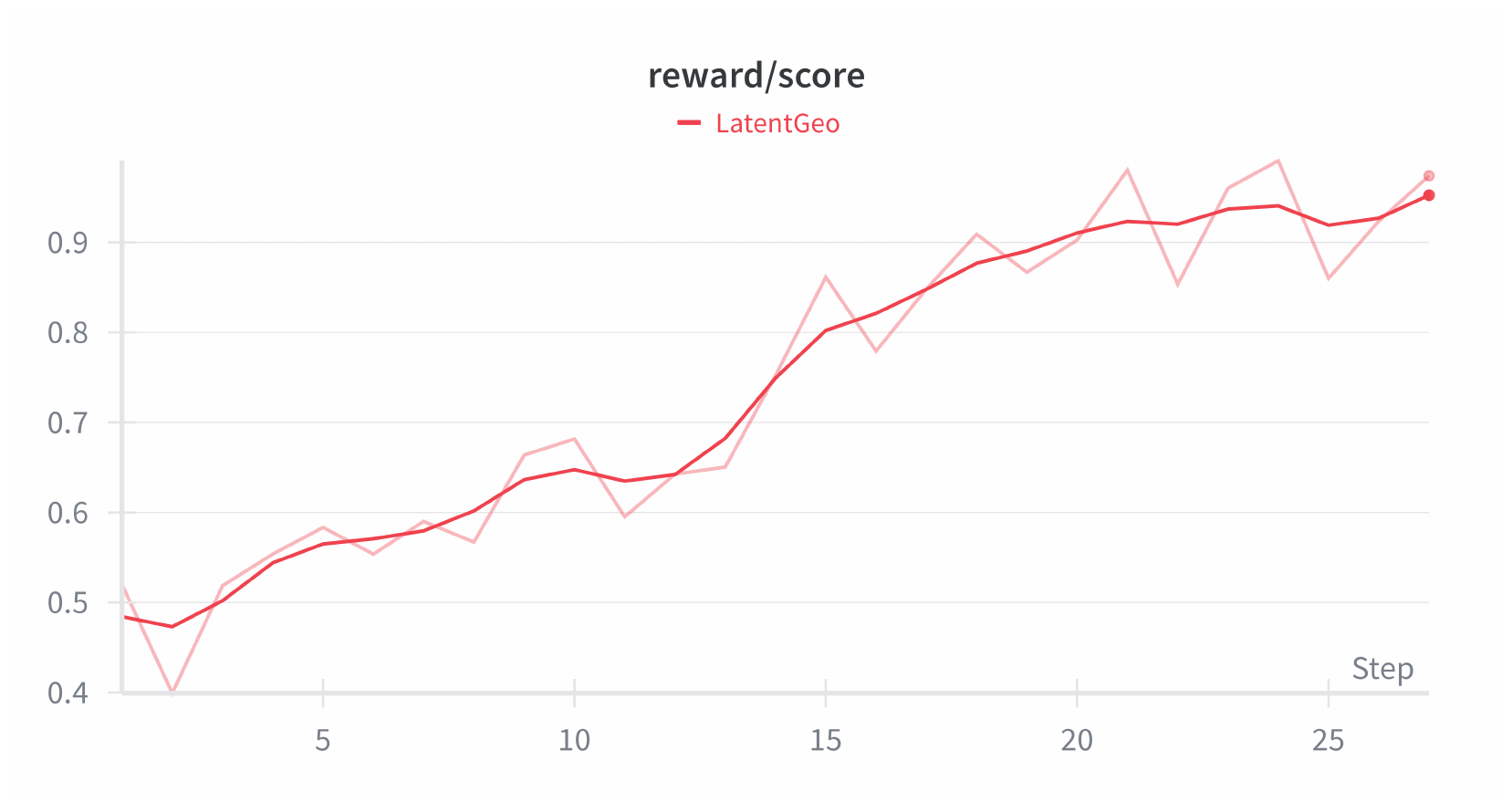}
    \caption{Overall trends}
    \label{fig:rl_all}
  \end{subfigure}\hfill
  \begin{subfigure}[t]{0.48\textwidth}
    \centering
    \includegraphics[width=\linewidth]{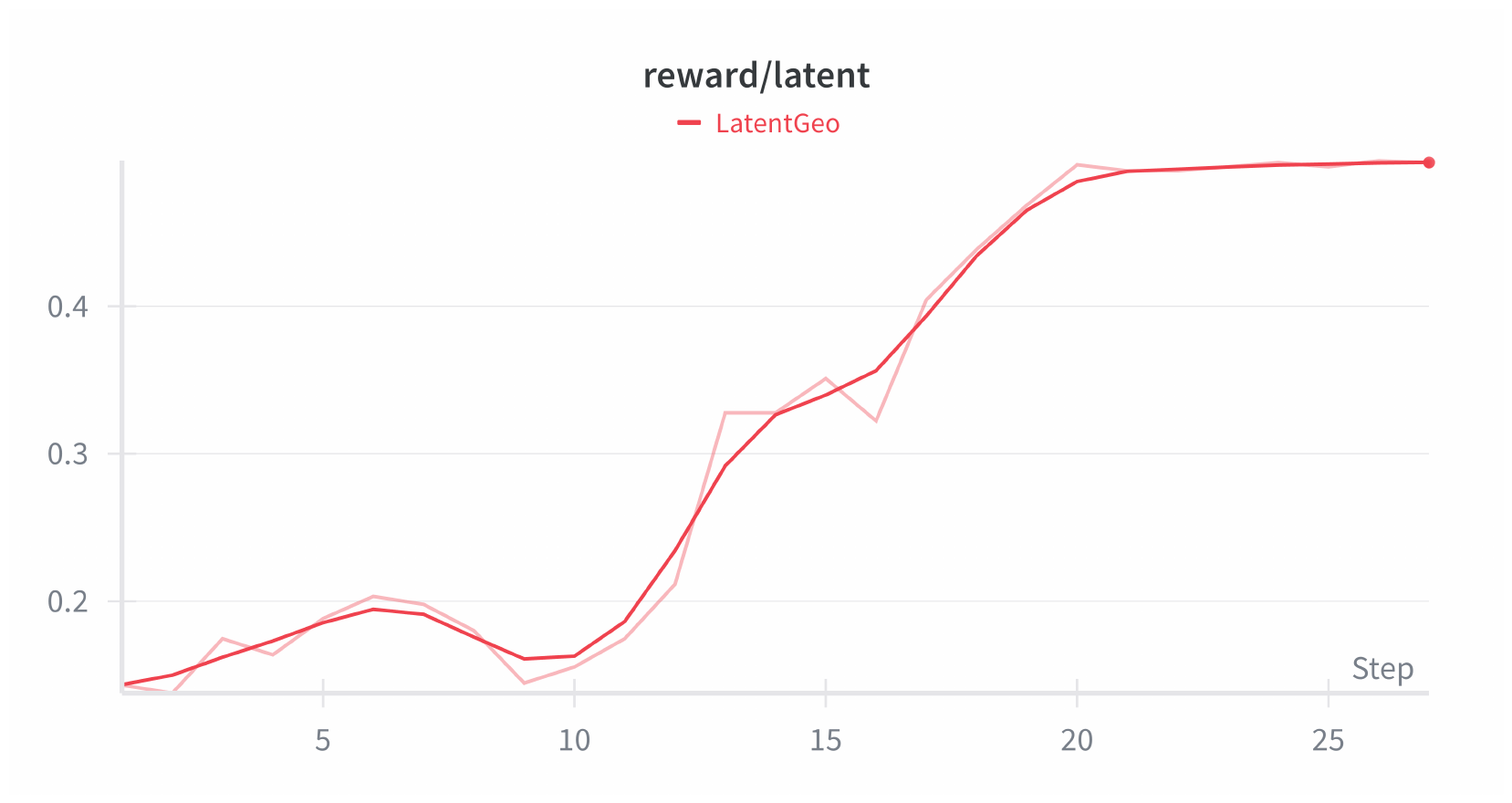}
    \caption{Latent-related trends}
    \label{fig:rl_latent}
  \end{subfigure}

  \vspace{0.5em}

  \begin{subfigure}[t]{0.48\textwidth}
    \centering
    \includegraphics[width=\linewidth]{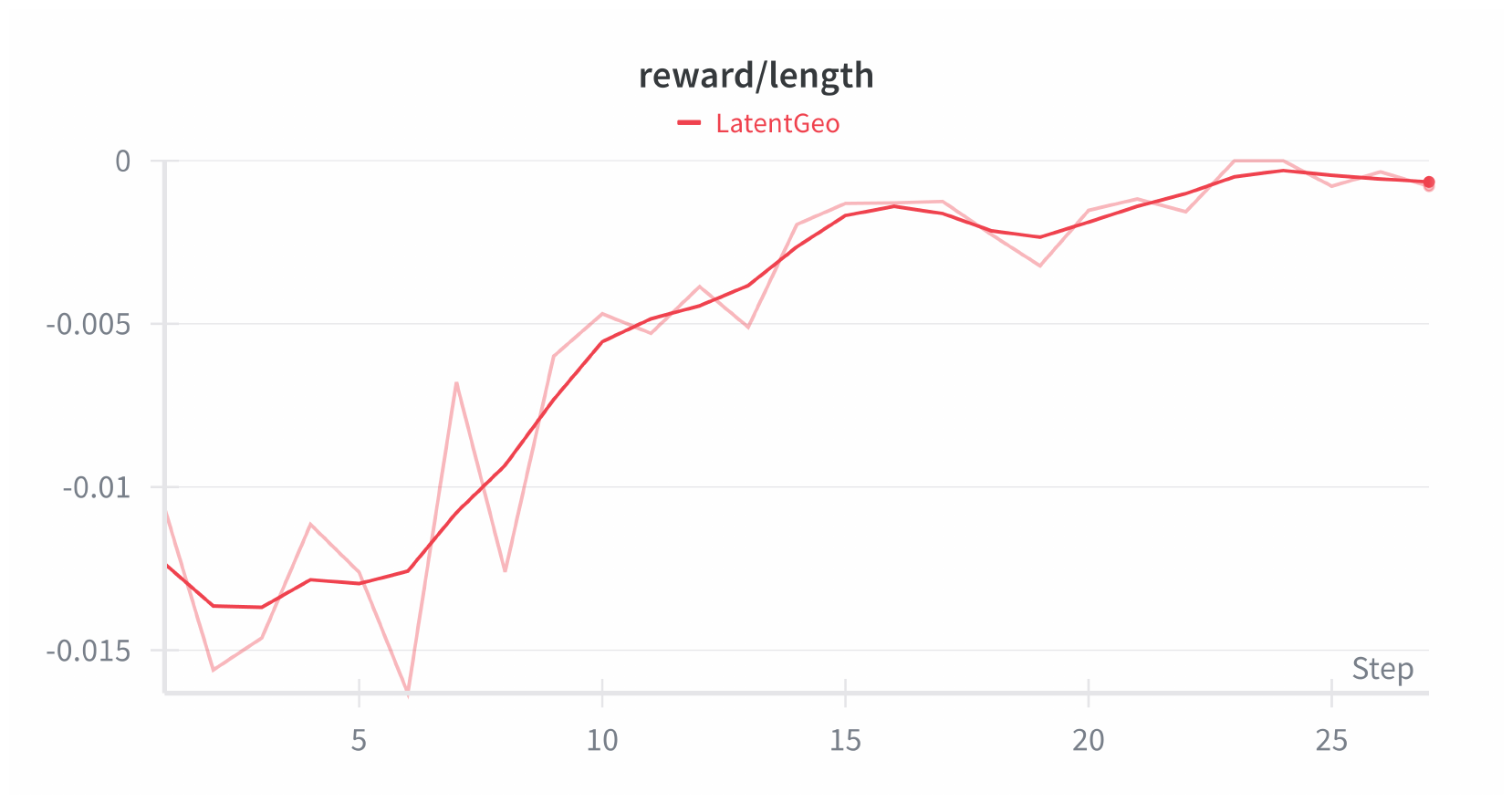}
    \caption{Length-related trends}
    \label{fig:rl_length}
  \end{subfigure}\hfill
  \begin{subfigure}[t]{0.48\textwidth}
    \centering
    \includegraphics[width=\linewidth]{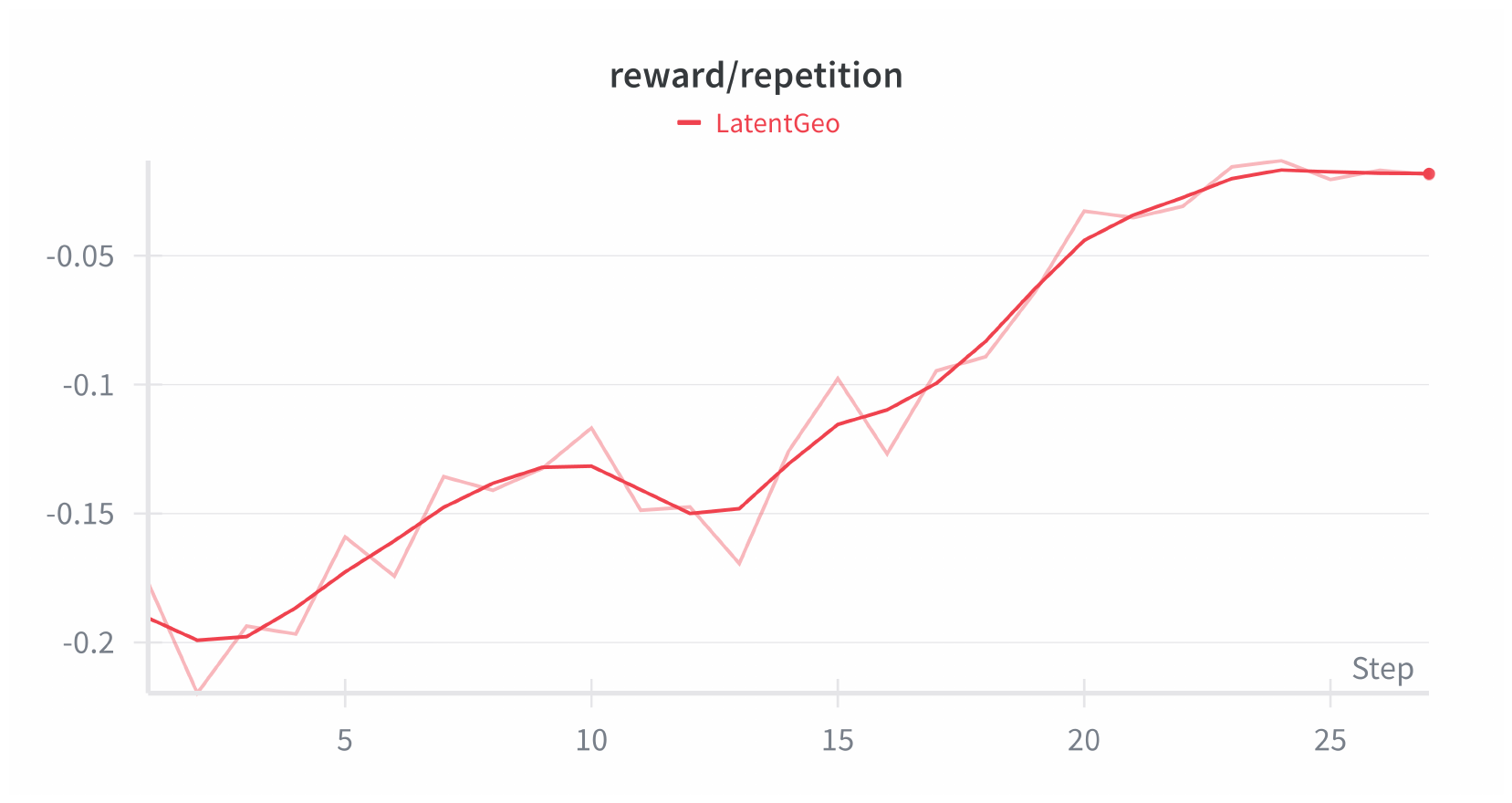}
    \caption{Repetition-related trends}
    \label{fig:rl_repetition}
  \end{subfigure}

  \caption{\textbf{RL training trends.} We report four perspectives of RL optimization: overall behavior, latent-token dynamics, length control, and repetition control.}
\end{figure}

% --------------------------------
% A.6: Baselines
% --------------------------------
\FloatBarrier
\subsection{Baselines}
\label{app:baseline}
%实验所用的模型介绍，也就是介绍主表中的模型+引用，在最后比较重要的是baseline要介绍清楚，加粗
We evaluate \textbf{LatentGeo} on \textsc{MATHVERSE} (\textit{testmini}) following the model selection in Table~\ref{tab:results_on_base}. We include Random and Human references, and three groups of model baselines. For \textit{Closed-Source MLLMs}, we include GPT-4o~\cite{openai2024gpt4o}, Gemini-1.5-Pro~\cite{team2023gemini}, and Qwen-VL-Plus~\cite{bai2023qwen}. For \textit{Open-Source General MLLMs}, we benchmark mPLUG-Owl2-7B~\cite{ye2023mplug}, LLaVA-1.5-13B~\cite{liu2024improved}, SPHINX-V2-13B~\cite{lin2023sphinx}, LLaVA-NeXT-34B~\cite{liu2024llava}, DeepSeek-VL~\cite{lu2024deepseek}, InternVL2-8B~\cite{chen2023internvl}, and Qwen2-VL~\cite{wang2024qwen2}. For \textit{Open-Source Math MLLMs}, we compare against Math-LLaVA-13B~\cite{shi2024math}, Math-PUMA-Qwen2-7B and Math-PUMA-DeepSeek-Math~\cite{zhuang2024math}, MAVIS-7B~\cite{zhang2025mavis}, InfiMM-Math~\cite{han24infimm}, and MultiMath-7B~\cite{peng2024multimath}.

We evaluate \textbf{LatentGeo} on the proposed \textsc{GeoAux} benchmark using two groups of baselines, as summarized in Table~\ref{tab:geoaux_main}. For \textit{Closed-Source MLLMs}, we include GPT-4o~\cite{openai2024gpt4o}, Qwen3-VL-Flash~\cite{bai2025qwen3vl}, and GPT-4o-mini~\cite{openai2024gpt4omini}. For \textit{Open-Source General MLLMs}, we benchmark LLaVA-1.6 (Vicuna-7B)~\cite{liu2024llava}, MMR1-7B-RL~\cite{leng2025mmr1}, OpenMMReasoner-RL~\cite{zhang2025openmmreasoner}, R1-Onevision-7B~\cite{yang2025r1onevision}, Qwen2.5-VL-7B and Qwen2.5-VL-32B~\cite{bai2025qwen25vl}, Qwen3-VL-8B~\cite{bai2025qwen3vl}, InternVL2-8B~\cite{chen2023internvl}, and InternVL3.5-8B~\cite{wang2025internvl35}. All models are evaluated under the same GeoAux protocol and report accuracy (\%) over 2,228 instances.

% --------------------------------
% A.7: Case Study
% --------------------------------
\FloatBarrier
\subsection{Case Study}
\label{app:case-study}

We present several qualitative inference cases from \textsc{GeoAux} to illustrate how \textbf{LatentGeo} introduces latent auxiliary constructions and completes the downstream geometric reasoning chain.

\begin{figure}[H]
\centering
\begin{suppexamplebox}{Representative Case Study Example}
\small
\suppfield{Knowledge.} Euclidean Plane Geometry.\\
\suppfield{Visual Operation.} Elemental Connectivity.

\vspace{1.5mm}
\suppfield{Question.} Hint: Please solve the problem and provide the final boxed answer as an exact angle measure, keeping degree notation if needed, e.g., \boxed{50^\circ}.\\
In the figure, $PA$ and $PB$ are tangents to circle $O$ at points $A$ and $B$, respectively. Point $E$ lies on circle $O$. If $\angle AEB=40^\circ$, find $\angle APB$.

\vspace{1.8mm}
\suppfield{Question Image.}
\begin{center}
\IfFileExists{figures/geoaux_19.pdf}{
  \includegraphics[width=0.34\linewidth]{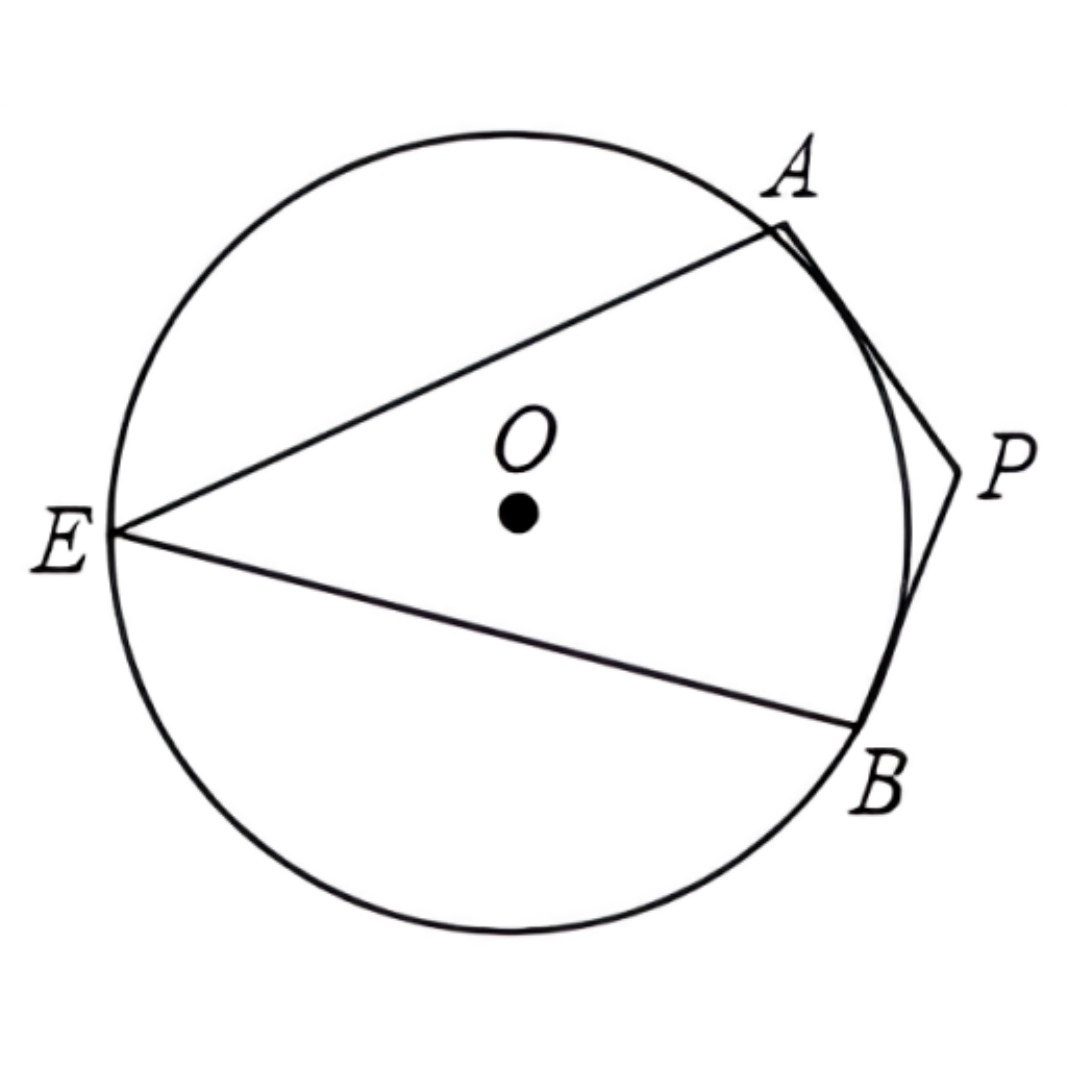}
}{
  \suppplaceholder{0.20\textheight}{}
}
\end{center}

\vspace{1.2mm}
\suppfield{Model Response.} Join $OA$ and $OB$. Since $PA$ and $PB$ are tangents to circle $O$, we have $OA\perp PA$ and $OB\perp PB$, so $\angle OAP=\angle OBP=90^\circ$.
\vspace{3mm}
\par\noindent\texttt{<|latent\_start|><|latent\_pad|>\dots<|latent\_pad|><|latent\_end|>}\par
\vspace{3mm}
The inscribed angle $\angle AEB$ subtends arc $AB$, so the central angle $\angle AOB$ subtending the same arc is twice the inscribed angle:
$\angle AOB=2\angle AEB=2\times 40^\circ=80^\circ$.
In quadrilateral $AOPB$, the angles at $A$ and $B$ are right angles, and the angle at $O$ is $80^\circ$, so the remaining angle at $P$ is
$\angle APB=360^\circ-90^\circ-90^\circ-80^\circ=100^\circ$.
Therefore, the answer is $\boxed{100}$.

\vspace{1.2mm}
\suppfield{Correct Answer.} $\boxed{100^\circ}$.

\vspace{1.2mm}
\end{suppexamplebox}
\caption{\textbf{Qualitative case study from \textsc{GeoAux}.} }
\label{fig:inference_case_template}
\end{figure}

\begin{figure}[H]
\centering
\begin{suppexamplebox}{Representative Case Study Example}
\small
\suppfield{Knowledge.} Euclidean Plane Geometry.\\
\suppfield{Visual Operation.} Orthogonal \& Parallel Alignment; Elemental Connectivity.

\vspace{1.5mm}
\suppfield{Question.} Hint: Please solve the problem and provide the final boxed answer as an exact value only, e.g., $\boxed{24}$.\\
In trapezoid $ABCD$, $EF$ is the midline. Segment $AF$ is drawn. If the area of $\triangle AEF$ is $6\,\text{cm}^2$, find the area of trapezoid $ABCD$ (in $\text{cm}^2$).

\vspace{1.8mm}
\suppfield{Question Image.}
\begin{center}
\IfFileExists{figures/geoaux_31.pdf}{
  \includegraphics[width=0.34\linewidth]{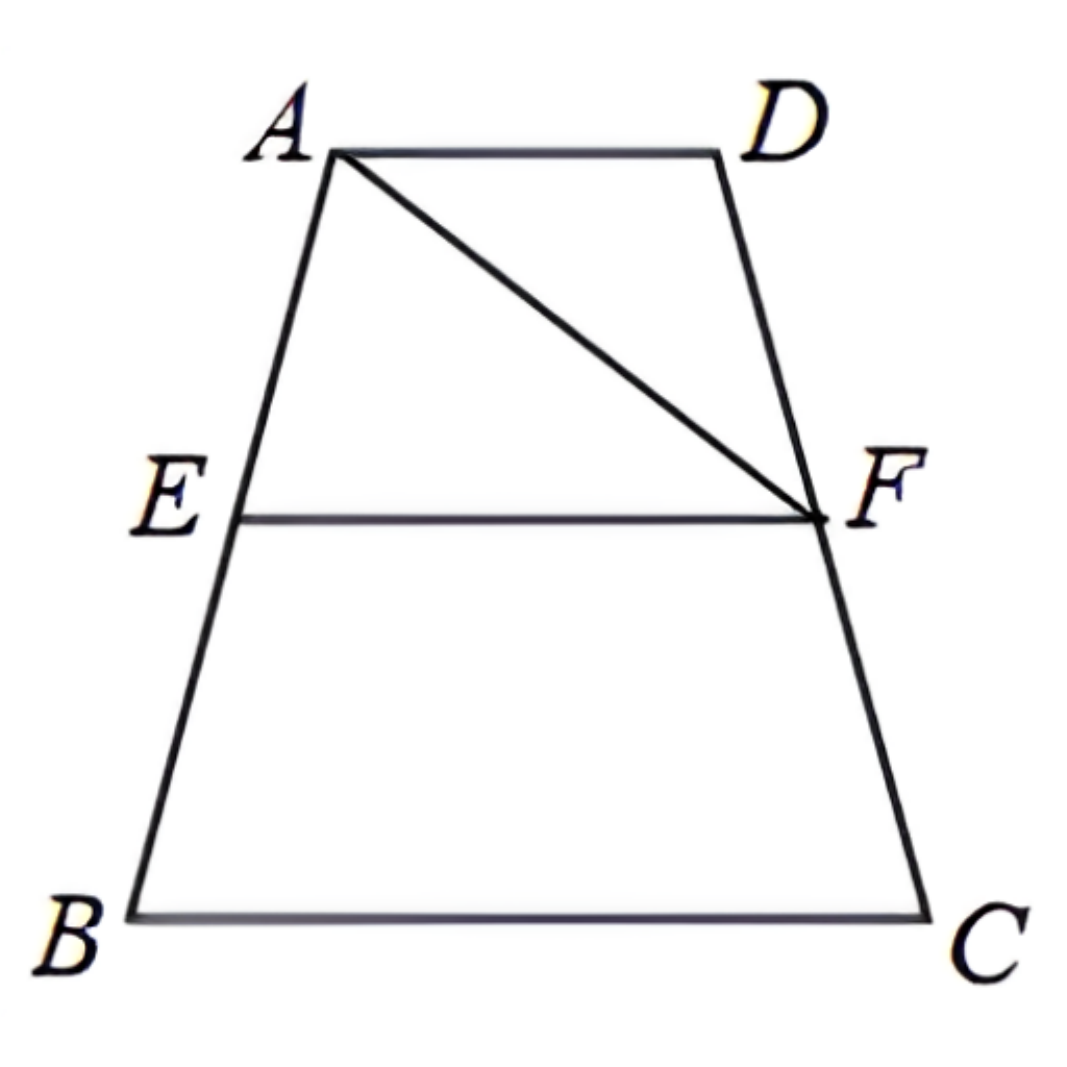}
}{
  \suppplaceholder{0.20\textheight}{}
}
\end{center}

\vspace{1.2mm}
\suppfield{Model Response.} Draw $AG \perp BC$ with $G$ on $BC$. Since $EF$ is the midline of trapezoid $ABCD$, the distance between the parallel sides $AB$ and $CD$ is $AG$, and the length of the midline is $EF = \tfrac{1}{2}(AB + CD)$. Therefore, the area of the trapezoid is $S_{ABCD} = \tfrac{1}{2}(AB + CD) \cdot AG = EF \cdot AG$.
\vspace{3mm}
\par\noindent\texttt{<|latent\_start|><|latent\_pad|>\dots<|latent\_pad|><|latent\_end|>}\par
\vspace{3mm}
Because $EF \parallel AB$, the triangles $\triangle AEF$ and $\triangle ABG$ are similar with similarity ratio $\tfrac{EF}{AB} = \tfrac{1}{2}$. Hence the area ratio is $\left(\tfrac{EF}{AB}\right)^2 = \tfrac{1}{4}$. Given $S_{\triangle AEF} = 6$, we have $S_{\triangle ABG} = 4 \times 6 = 24$. Since $S_{ABCD} = S_{\triangle ABG}$ (the area of the trapezoid equals the area of the larger similar triangle with the same height $AG$), the area of trapezoid $ABCD$ is $24$. Therefore, the answer is $\boxed{24}$.

\vspace{1.2mm}
\suppfield{Correct Answer.} $\boxed{24}$.

\vspace{1.2mm}
\end{suppexamplebox}
\caption{\textbf{Qualitative case study from \textsc{GeoAux}.} }
\label{fig:inference_case_geoaux31}
\end{figure}

\begin{figure}[H]
\centering
\begin{suppexamplebox}{Representative Case Study Example}
\small
\suppfield{Knowledge.} Spatial \& Projective Geometry.\\
\suppfield{Visual Operation.} Spatial Projection \& Unfolding; Elemental Connectivity.

\vspace{1.5mm}
\suppfield{Question.} Hint: Please solve the problem and provide the final boxed answer as a simplified expression only, e.g., $\boxed{40\sqrt{5}}$.\\
A cube gift box has side length 20 cm. A gecko is at vertex $A$ on the bottom face, and a mosquito is at vertex $C'$ on the top face, as shown. The gecko must reach the mosquito within 0.5 minutes, moving along the surfaces of the box. What is the minimum speed the gecko must crawl per minute (in cm/min)? Express your answer in simplest radical form.

\vspace{1.8mm}
\suppfield{Question Image.}
\begin{center}
\IfFileExists{figures/geoaux_1021.pdf}{
  \includegraphics[width=0.34\linewidth]{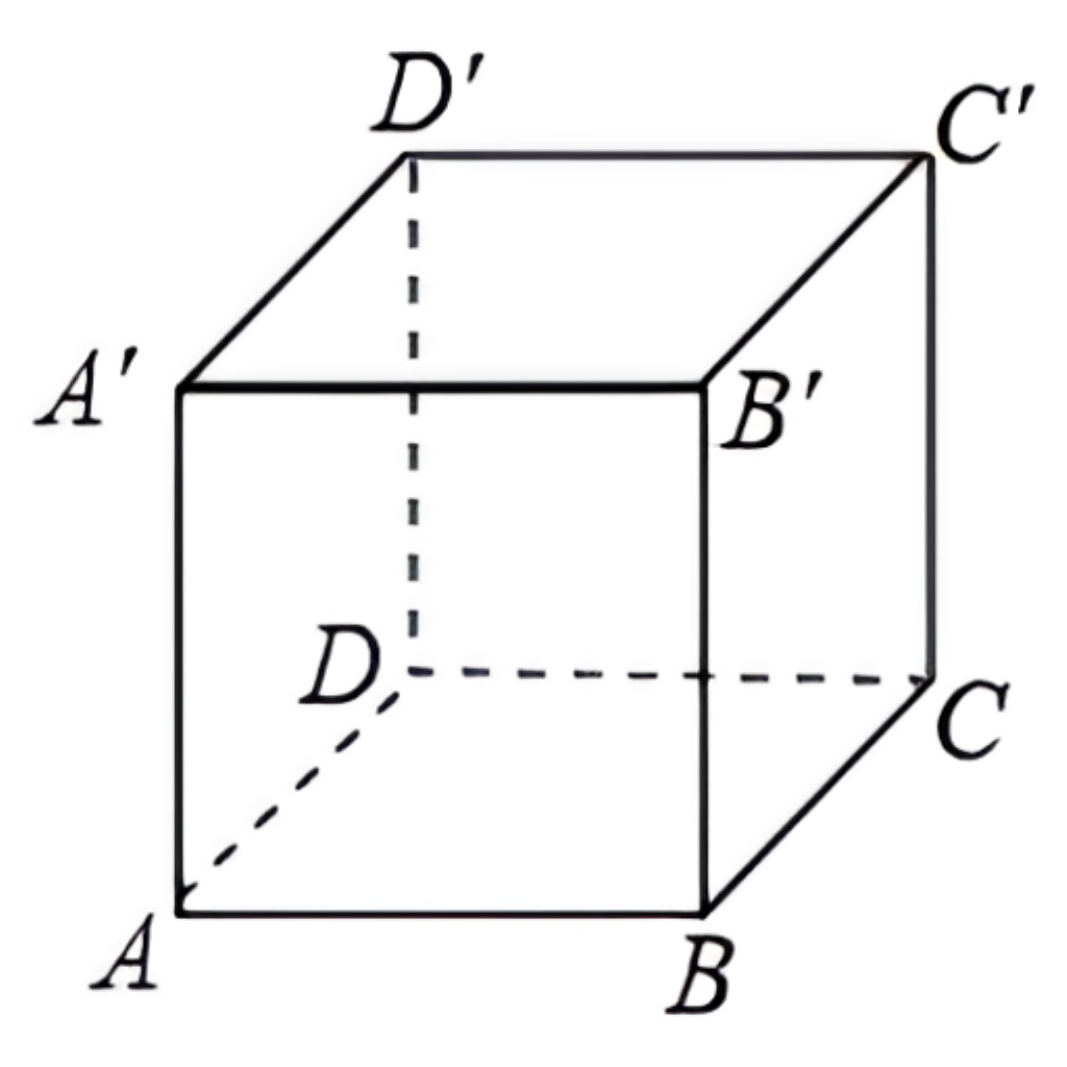}
}{
  \suppplaceholder{0.20\textheight}{}
}
\end{center}

\vspace{1.2mm}
\suppfield{Model Response.} Consider the net of the cube that unfolds the two faces containing the gecko and the mosquito so that the shortest path on the surfaces becomes a straight line in the net. In the net, the gecko starts at $A$ on the bottom face and the mosquito is at $C'$ on the top face. The straight-line distance between $A$ and $C'$ in the net is the hypotenuse of a right triangle with legs $20$ and $40$ (since the bottom and top faces are separated by $20$ cm and the horizontal distance between the corresponding positions on the two faces is $40$ cm).
\vspace{3mm}
\par\noindent\texttt{<|latent\_start|><|latent\_pad|>\dots<|latent\_pad|><|latent\_end|>}\par
\vspace{3mm}
Thus the shortest path length is $\sqrt{20^2 + 40^2} = \sqrt{400 + 1600} = \sqrt{2000} = 20\sqrt{5}$ cm. The gecko has 0.5 minutes to cover this distance, so the required minimum speed is $\dfrac{20\sqrt{5}}{0.5} = 40\sqrt{5}$ cm/min. Therefore, the answer is $\boxed{40\sqrt{5}}$.

\vspace{1.2mm}
\suppfield{Correct Answer.} $\boxed{40\sqrt{5}}$.

\vspace{1.2mm}
\end{suppexamplebox}
\caption{\textbf{Qualitative case study from \textsc{GeoAux}.} }
\label{fig:inference_case_geoaux1021}
\end{figure}

\begin{figure}[H]
\centering
\begin{suppexamplebox}{Representative Case Study Example}
\small
\suppfield{Knowledge.} Analytic Coordinate Geometry.\\
\suppfield{Visual Operation.} Analytic Overlay.

\vspace{1.5mm}
\suppfield{Question.} Hint: Please solve the problem and provide the final boxed answer as the uppercase option letter only, e.g., $\boxed{A}$.\\
In parallelogram $ABCD$, three vertices are $A(-1,0)$, $B(-2,-3)$, and $C(2,-1)$. What are the coordinates of $D$?\\
Options: A. $(3,1)$ \quad B. $(3,2)$ \quad C. $(3,3)$ \quad D. $(3,4)$

\vspace{1.8mm}
\suppfield{Question Image.}
\begin{center}
\IfFileExists{figures/geoaux_1389.pdf}{
  \includegraphics[width=0.34\linewidth]{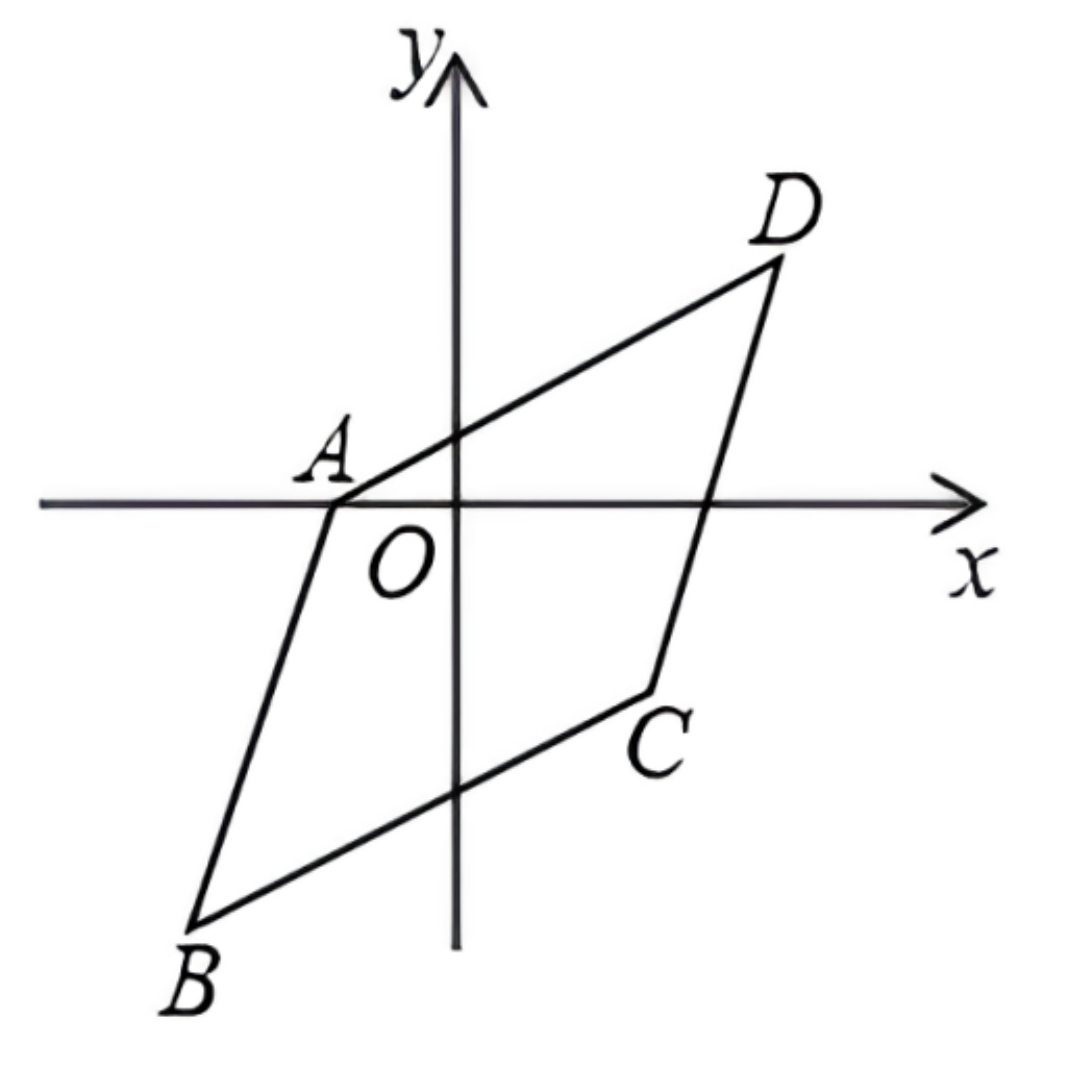}
}{
  \suppplaceholder{0.20\textheight}{}
}
\end{center}

\vspace{1.2mm}
\suppfield{Model Response.} Since $ABCD$ is a parallelogram, the diagonals $AC$ and $BD$ bisect each other. Therefore, the midpoint of $AC$ equals the midpoint of $BD$.
\vspace{3mm}
\par\noindent\texttt{<|latent\_start|><|latent\_pad|>\dots<|latent\_pad|><|latent\_end|>}\par
\vspace{3mm}
Compute the midpoint of $AC$: the midpoint formula gives $\left(\frac{-1+2}{2},\frac{0+(-1)}{2}\right)=\left(\frac{1}{2},-\frac{1}{2}\right)$. Let $D=(x,y)$. Then the midpoint of $BD$ is $\left(\frac{-2+x}{2},\frac{-3+y}{2}\right)$. Setting the two midpoints equal gives the system $\displaystyle \frac{-2+x}{2}=\frac{1}{2}$ and $\displaystyle \frac{-3+y}{2}=-\frac{1}{2}$. Solve the first: $-2+x=1\Rightarrow x=3$. Solve the second: $-3+y=-1\Rightarrow y=2$. Hence $D=(3,2)$. Therefore, the correct option is $\boxed{B}$.

\vspace{1.2mm}
\suppfield{Correct Answer.} $\boxed{B}$.

\vspace{1.2mm}
\end{suppexamplebox}
\caption{\textbf{Qualitative case study from \textsc{GeoAux}.} }
\label{fig:inference_case_geoaux1389}
\end{figure}

\begin{figure}[H]
\centering
\begin{suppexamplebox}{Representative Case Study Example}
\small
\suppfield{Knowledge.} Functional Graph Geometry.\\
\suppfield{Visual Operation.} Analytic Overlay.

\vspace{1.5mm}
\suppfield{Question.} Hint: Please solve the problem and provide the final boxed answer as the uppercase option letter only, e.g., $\boxed{A}$.\\
The arch of a bridge is parabolic. When the arch apex is 2 m above the water surface, the water surface width is 4 m. If the water surface drops by 1 m, by how much does the water surface width increase?\\
Options: A. $1\,m$ \quad B. $2\,m$ \quad C. $(2\sqrt{6}-4)\,m$ \quad D. $(\sqrt{6}-2)\,m$

\vspace{1.8mm}
\suppfield{Question Image.}
\begin{center}
\IfFileExists{figures/geoaux_1618.pdf}{
  \includegraphics[width=0.34\linewidth]{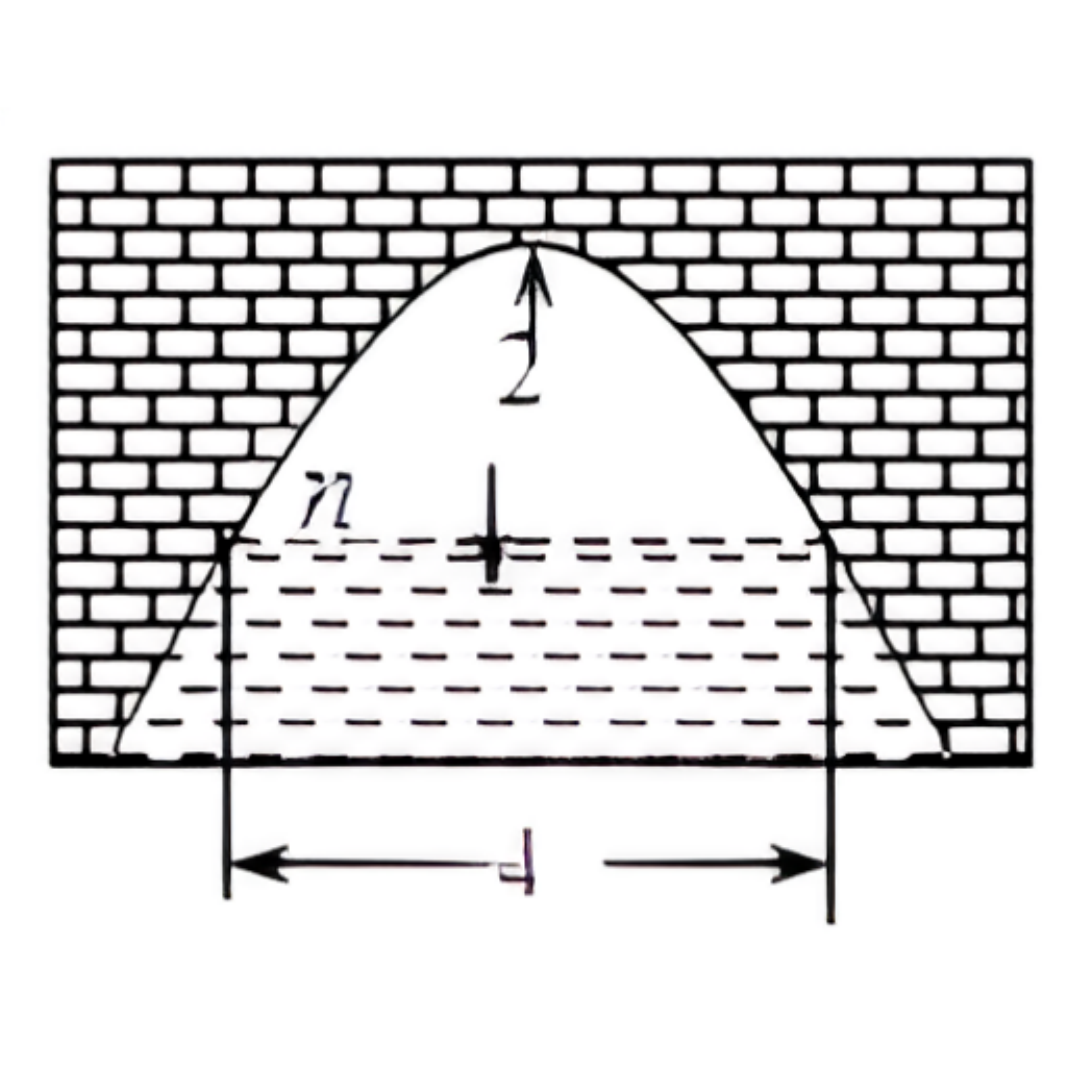}
}{
  \suppplaceholder{0.20\textheight}{}
}
\end{center}

\vspace{1.2mm}
\suppfield{Model Response.} Set up a coordinate system with the vertex of the parabola at the origin and the axis of symmetry along the $y$-axis. Then the parabola is $y=ax^2$ with $a<0$. The water surface at the initial level is the line $y=-2$, and the water surface at the final level is the line $y=-3$.
\vspace{3mm}
\par\noindent\texttt{<|latent\_start|><|latent\_pad|>\dots<|latent\_pad|><|latent\_end|>}\par
\vspace{3mm}
At the initial level $y=-2$, the water surface is the line $y=-2$. The intersection with the parabola is at $y=-2$, so $-2=ax^2$, giving $x=\pm\sqrt{-\frac{2}{a}}$. The width is $2\sqrt{-\frac{2}{a}}$. At the final level $y=-3$, the intersection is at $y=-3$, so $-3=ax^2$, giving $x=\pm\sqrt{-\frac{3}{a}}$. The new width is $2\sqrt{-\frac{3}{a}}$. The increase in width is $\displaystyle 2\sqrt{-\frac{3}{a}}-2\sqrt{-\frac{2}{a}}=2\sqrt{-\frac{1}{a}}\Big(\sqrt{3}-\sqrt{2}\Big)=2\sqrt{6}-4.$ Therefore, the correct option is $\boxed{C}$.

\vspace{1.2mm}
\suppfield{Correct Answer.} $\boxed{C}$.

\vspace{1.2mm}
\end{suppexamplebox}
\caption{\textbf{Qualitative case study from \textsc{GeoAux}.} }
\label{fig:inference_case_geoaux1618}
\end{figure}

%% file: main.bib
@String(CVPR= {IEEE Conf. Comput. Vis. Pattern Recog. (CVPR)})

@String(ECCV= {Eur. Conf. Comput. Vis. (ECCV)})

@String(NIPS= {Adv. Neural Inform. Process. Syst. (NeurIPS)})

@String(ICLR = {Int. Conf. Learn. Represent. (ICLR)})

@String(ACL = {Ann. Meet. Assoc. Comput. Linguist. (ACL)})

@inproceedings{lu2024mathvista,
  title     = {MathVista: Evaluating Mathematical Reasoning of Foundation Models in Visual Contexts},
  author    = {Lu, Pan and Bansal, Hritik and Xia, Tony and Liu, Jiacheng and Li, Chunyuan and Hajishirzi, Hannaneh and Cheng, Hao and Chang, Kai-Wei and Galley, Michel and Gao, Jianfeng},
  booktitle = ICLR,
  year      = {2024},
  url       = {https://openreview.net/forum?id=KUNzEQMWU7}
}

@inproceedings{zhang2024mathverse,
  title     = {MathVerse: Does Your Multi-modal LLM Truly See the Diagrams in Visual Math Problems?},
  author    = {Zhang, Renrui and Jiang, Dongzhi and Zhang, Yichi and Lin, Haokun and Guo, Ziyu and Qiu, Pengshuo and Zhou, Aojun and Lu, Pan and Chang, Kai-Wei and Gao, Peng and Li, Hongsheng},
  booktitle = ECCV,
  year      = {2024},
  url       = {https://arxiv.org/abs/2403.14624}
}

@inproceedings{wang2024mathvision,
  title     = {Measuring Multimodal Mathematical Reasoning with {MATH-Vision}},
  author    = {Wang, Ke and Pan, Junting and Shi, Weikang and Lu, Zimu and Ren, Houxing and Zhou, Aojun and Zhan, Mingjie and Li, Hongsheng},
  booktitle = NIPS,
  note      = {Datasets and Benchmarks Track},
  year      = {2024},
  url       = {https://proceedings.neurips.cc/paper_files/paper/2024/hash/ad0edc7d5fa1a783f063646968b7315b-Abstract-Datasets_and_Benchmarks_Track.html}
}

@inproceedings{wang2025mvmath,
  title     = {{MV-MATH}: Evaluating Multimodal Math Reasoning in Multi-Visual Contexts},
  author    = {Wang, Peijie and Li, Zhong-Zhi and Yin, Fei and Yang, Xin and Ran, Dekang and Liu, Cheng-Lin},
  booktitle = CVPR,
  year      = {2025},
  url       = {https://arxiv.org/abs/2502.20808}
}

@misc{wang2025solidgeo,
  title        = {{SOLIDGEO}: Measuring Multimodal Spatial Math Reasoning in Solid Geometry},
  author       = {Wang, Peijie and Yang, Chao and Li, Zhong-Zhi and Yin, Fei and Ran, Dekang and Tian, Mi and Ji, Zhilong and Bai, Jinfeng and Liu, Cheng-Lin},
  year         = {2025},
  eprint       = {2505.21177},
  archivePrefix= {arXiv},
  url          = {https://arxiv.org/abs/2505.21177}
}

@inproceedings{wang2025mathcodervl,
  title     = {MathCoder-VL: Bridging Vision and Code for Enhanced Multimodal Mathematical Reasoning},
  author    = {Wang, Ke and Pan, Junting and Wei, Linda and Zhou, Aojun and Shi, Weikang and Lu, Zimu and Xiao, Han and Yang, Yunqiao and Ren, Houxing and Zhan, Mingjie and Li, Hongsheng},
  booktitle = {Findings of the Association for Computational Linguistics (ACL)},
  year      = {2025},
  url       = {https://aclanthology.org/2025.findings-acl.128/}
}

@inproceedings{wu2024egps,
  title     = {{E-GPS}: Explainable Geometry Problem Solving via Top-Down Solver and Bottom-Up Generator},
  author    = {Wu, Wenjun and Zhang, Lingling and Liu, Jun and Tang, Xi and Wang, Yaxian and Wang, Shaowei and Wang, Qianying},
  booktitle = CVPR,
  year      = {2024},
  url       = {https://openaccess.thecvf.com/content/CVPR2024/papers/Wu_E-GPS_Explainable_Geometry_Problem_Solving_via_Top-Down_Solver_and_Bottom-Up_CVPR_2024_paper.pdf}
}

@article{wei2025geointr1,
  title   = {Geoint-R1: Formalizing Multimodal Geometric Reasoning with Dynamic Auxiliary Constructions},
  author  = {Wei, Jingxuan and Jia, Caijun and Chen, Qi and He, Honghao and Sun, Linzhuang and He, Conghui and Wu, Lijun and Yu, Bihui and Tan, Cheng},
  journal = {arXiv preprint arXiv:2508.03173},
  year    = {2025},
  url     = {https://arxiv.org/abs/2508.03173}
}

@article{wang2025geometryzero,
  title   = {GeometryZero: Improving Geometry Solving for LLM with Group Contrastive Policy Optimization},
  author  = {Wang, Yikun and Wang, Yibin and Wang, Dianyi and Peng, Zimian and Guo, Qipeng and Tao, Dacheng and Wang, Jiaqi},
  journal = {arXiv preprint arXiv:2506.07160},
  year    = {2025},
  url     = {https://arxiv.org/abs/2506.07160}
}

@inproceedings{hu2024visualsketchpad,
  title     = {Visual Sketchpad: Sketching as a Visual Chain of Thought for Multimodal Language Models},
  author    = {Hu, Yushi and Shi, Weijia and Fu, Xingyu and Roth, Dan and Ostendorf, Mari and Zettlemoyer, Luke and Smith, Noah A. and Krishna, Ranjay},
  booktitle = NIPS,
  year      = {2024},
  url       = {https://arxiv.org/abs/2406.09403}
}

@inproceedings{zhang2025mavis,
  title     = {{MAVIS}: Mathematical Visual Instruction Tuning with an Automatic Data Engine},
  author    = {Zhang, Renrui and Wei, Xinyu and Jiang, Dongzhi and Guo, Ziyu and Li, Shicheng and Zhang, Yichi and Tong, Chengzhuo and Liu, Jiaming and Zhou, Aojun and Wei, Bin and Zhang, Shanghang and Gao, Peng and Li, Chunyuan and Li, Hongsheng},
  booktitle = ICLR,
  year      = {2025},
  url       = {https://arxiv.org/abs/2407.08739}
}

@misc{chung2025v1revisitation,
  title         = {v1: Learning to Point Visual Tokens for Multimodal Grounded Reasoning},
  author        = {Chung, Jiwan and Kim, Junhyeok and Kim, Siyeol and Lee, Jaeyoung and Kim, Min Soo and Yu, Youngjae},
  year          = {2025},
  eprint        = {2505.18842},
  archivePrefix = {arXiv},
  primaryClass  = {cs.CL},
  url           = {https://arxiv.org/abs/2505.18842}
}

@article{zhang2025openeyes,
  title   = {Open Eyes, Then Reason: Fine-grained Visual Mathematical Understanding in Multimodal Large Language Models},
  author  = {Zhang, Shan and Chen, Aotian and Sun, Yanpeng and Gu, Jindong and Zheng, Yi-Yu and Koniusz, Piotr and Zou, Kai and van den Hengel, Anton and Xue, Yuan},
  journal = {arXiv preprint arXiv:2501.06430},
  year    = {2025},
  url     = {https://arxiv.org/abs/2501.06430}
}

@misc{sun2025lacot_visual,
  title         = {Latent Chain-of-Thought for Visual Reasoning},
  author        = {Guohao Sun and Hang Hua and Jian Wang and Jiebo Luo and Sohail Dianat and Majid Rabbani and Raghuveer Rao and Zhiqiang Tao},
  year          = {2025},
  eprint        = {2510.23925},
  archivePrefix = {arXiv},
  primaryClass  = {cs.AI},
  url           = {https://arxiv.org/abs/2510.23925}
}

@misc{yang2025machinementalimageryempower,
      title={Machine Mental Imagery: Empower Multimodal Reasoning with Latent Visual Tokens}, 
      author={Zeyuan Yang and Xueyang Yu and Delin Chen and Maohao Shen and Chuang Gan},
      year={2025},
      eprint={2506.17218},
      archivePrefix={arXiv},
      primaryClass={cs.CV},
      url={https://arxiv.org/abs/2506.17218}, 
}

@misc{lvr2025,
      title={Latent Visual Reasoning}, 
      author={Bangzheng Li and Ximeng Sun and Jiang Liu and Ze Wang and Jialian Wu and Xiaodong Yu and Hao Chen and Emad Barsoum and Muhao Chen and Zicheng Liu},
      year={2025},
      eprint={2509.24251},
      archivePrefix={arXiv},
      primaryClass={cs.CV},
      url={https://arxiv.org/abs/2509.24251}, 
}

@misc{qin2025chainofvisualthought,
      title={Chain-of-Visual-Thought: Teaching VLMs to See and Think Better with Continuous Visual Tokens}, 
      author={Yiming Qin and Bomin Wei and Jiaxin Ge and Konstantinos Kallidromitis and Stephanie Fu and Trevor Darrell and Xudong Wang},
      year={2025},
      eprint={2511.19418},
      archivePrefix={arXiv},
      primaryClass={cs.CV},
      url={https://arxiv.org/abs/2511.19418}, 
}

@misc{zhang2025latentsketchpad,
  title         = {Latent Sketchpad: Sketching Visual Thoughts to Elicit Multimodal Reasoning in MLLMs},
  author        = {Zhang, Huanyu and Wu, Wenshan and Li, Chengzu and Shang, Ning and Xia, Yan and Huang, Yangyu and Zhang, Yifan and Dong, Li and Zhang, Zhang and Wang, Liang and Tan, Tieniu and Wei, Furu},
  year          = {2025},
  eprint        = {2510.24514},
  archivePrefix = {arXiv},
  primaryClass  = {cs.CV},
  doi           = {10.48550/arXiv.2510.24514},
  url           = {https://arxiv.org/abs/2510.24514}
}

@misc{wang2025monetreasoninglatentvisual,
      title={Monet: Reasoning in Latent Visual Space Beyond Images and Language}, 
      author={Qixun Wang and Yang Shi and Yifei Wang and Yuanxing Zhang and Pengfei Wan and Kun Gai and Xianghua Ying and Yisen Wang},
      year={2025},
      eprint={2511.21395},
      archivePrefix={arXiv},
      primaryClass={cs.CV},
      url={https://arxiv.org/abs/2511.21395}, 
}

@misc{openai2024gpt4o,
  title={{GPT-4o} System Card},
  author={OpenAI},
  year={2024},
  url={https://openai.com/research/gpt-4o-system-card}
}

@article{team2024gemini15,
  title   = {Gemini 1.5: Unlocking multimodal understanding across millions of tokens of context},
  author  = {{Gemini Team Google} and Georgiev, Petko and Lei, Ving Ian and Burnell, Ryan and Bai, Libin and Gulati, Anmol and others},
  journal = {arXiv preprint arXiv:2403.05530},
  year    = {2024},
  doi     = {10.48550/arXiv.2403.05530},
  url     = {https://arxiv.org/abs/2403.05530}
}

@misc{bai2023qwen,
      title={Qwen-VL: A Versatile Vision-Language Model for Understanding, Localization, Text Reading, and Beyond}, 
      author={Jinze Bai and Shuai Bai and Shusheng Yang and Shijie Wang and Sinan Tan and Peng Wang and Junyang Lin and Chang Zhou and Jingren Zhou},
      year={2023},
      eprint={2308.12966},
      archivePrefix={arXiv},
      primaryClass={cs.CV},
      url={https://arxiv.org/abs/2308.12966}, 
}

@article{ye2023mplug,
  title={mplug-owl: Modularization empowers large language models with multimodality},
  author={Ye, Qinghao and Xu, Haiyang and Xu, Guohai and Ye, Jiabo and Yan, Ming and Zhou, Yiyang and Wang, Junyang and Hu, Anwen and Shi, Pengcheng and Shi, Yaya and others},
  journal={arXiv preprint arXiv:2304.14178},
  year={2023},
  url={https://arxiv.org/abs/2304.14178}
}

@inproceedings{liu2024improved,
  title={Improved baselines with visual instruction tuning},
  author={Liu, Haotian and Li, Chunyuan and Li, Yuheng and Lee, Yong Jae},
  booktitle=CVPR,
  pages={26296--26306},
  year={2024},
  url={https://arxiv.org/abs/2310.03744}
}

@misc{lin2023sphinx,
      title={SPHINX: The Joint Mixing of Weights, Tasks, and Visual Embeddings for Multi-modal Large Language Models}, 
      author={Ziyi Lin and Chris Liu and Renrui Zhang and Peng Gao and Longtian Qiu and Han Xiao and Han Qiu and Chen Lin and Wenqi Shao and Keqin Chen and Jiaming Han and Siyuan Huang and Yichi Zhang and Xuming He and Hongsheng Li and Yu Qiao},
      year={2023},
      eprint={2311.07575},
      archivePrefix={arXiv},
      primaryClass={cs.CV},
      url={https://arxiv.org/abs/2311.07575}, 
}

@misc{liu2024llava,
  title        = {LLaVA-NeXT: Improved reasoning, OCR, and world knowledge},
  author       = {Liu, Haotian and Li, Chunyuan and Li, Yuheng and Li, Bo and Zhang, Yuanhan and Shen, Sheng and Lee, Yong Jae},
  year         = {2024},
  url          = {https://llava-vl.github.io/blog/2024-01-30-llava-next/},
}

@misc{leng2025mmr1,
  title        = {MMR1: Enhancing Multimodal Reasoning with Variance-Aware Sampling and Open Resources},
  author       = {Leng, Sicong and Wang, Jing and Li, Jiaxi and Zhang, Hao and Hu, Zhiqiang and Zhang, Boqiang and Jiang, Yuming and Zhang, Hang and Li, Xin and Bing, Lidong and Zhao, Deli and Lu, Wei and Rong, Yu and Sun, Aixin and Lu, Shijian},
  year         = {2025},
  eprint       = {2509.21268},
  archivePrefix= {arXiv},
  primaryClass = {cs.CV},
  url          = {https://arxiv.org/abs/2509.21268},
}

@misc{zhang2025openmmreasoner,
  title        = {OpenMMReasoner: Pushing the Frontiers for Multimodal Reasoning with an Open and General Recipe},
  author       = {Zhang, Kaichen and Wu, Keming and Yang, Zuhao and Li, Bo and Hu, Kairui and Wang, Bin and Liu, Ziwei and Li, Xingxuan and Bing, Lidong},
  year         = {2025},
  eprint       = {2511.16334},
  archivePrefix= {arXiv},
  primaryClass = {cs.AI},
  url          = {https://arxiv.org/abs/2511.16334},
}

@misc{yang2025r1onevision,
  title        = {R1-Onevision: Advancing Generalized Multimodal Reasoning through Cross-Modal Formalization},
  author       = {Yang, Yi and He, Xiaoxuan and Pan, Hongkun and Jiang, Xiyan and Deng, Yan and Yang, Xingtao and Lu, Haoyu and Yin, Dacheng and Rao, Fengyun and Zhu, Minfeng and Zhang, Bo and Chen, Wei},
  year         = {2025},
  eprint       = {2503.10615},
  archivePrefix= {arXiv},
  primaryClass = {cs.CV},
  doi          = {10.48550/arXiv.2503.10615},
  url          = {https://arxiv.org/abs/2503.10615},

}

@misc{lu2024deepseek,
      title={DeepSeek-VL: Towards Real-World Vision-Language Understanding}, 
      author={Haoyu Lu and Wen Liu and Bo Zhang and Bingxuan Wang and Kai Dong and Bo Liu and Jingxiang Sun and Tongzheng Ren and Zhuoshu Li and Hao Yang and Yaofeng Sun and Chengqi Deng and Hanwei Xu and Zhenda Xie and Chong Ruan},
      year={2024},
      eprint={2403.05525},
      archivePrefix={arXiv},
      primaryClass={cs.AI},
      url={https://arxiv.org/abs/2403.05525}, 
}

@misc{shao2024deepseekmath,
  title        = {DeepSeekMath: Pushing the Limits of Mathematical Reasoning in Open Language Models},
  author       = {Shao, Zhihong and Wang, Peiyi and Zhu, Qihao and Xu, Runxin and Song, Junxiao and Bi, Xiao and Zhang, Haowei and Zhang, Mingchuan and Li, Y. K. and Wu, Y. and Guo, Daya},
  year         = {2024},
  eprint       = {2402.03300},
  archivePrefix= {arXiv},
  primaryClass = {cs.CL},
  doi          = {10.48550/arXiv.2402.03300},
  url          = {https://arxiv.org/abs/2402.03300}
}

@misc{chen2023internvl,
      title={InternVL: Scaling up Vision Foundation Models and Aligning for Generic Visual-Linguistic Tasks}, 
      author={Zhe Chen and Jiannan Wu and Wenhai Wang and Weijie Su and Guo Chen and Sen Xing and Muyan Zhong and Qinglong Zhang and Xizhou Zhu and Lewei Lu and Bin Li and Ping Luo and Tong Lu and Yu Qiao and Jifeng Dai},
      year={2023},
      eprint={2312.14238},
      archivePrefix={arXiv},
      primaryClass={cs.CV},
      url={https://arxiv.org/abs/2312.14238}, 
}

@article{shi2024math,
  title={Math-llava: Bootstrapping mathematical reasoning for multimodal large language models},
  author={Shi, Wenhao and Hu, Zhiqiang and Bin, Yi and Liu, Junhua and Yang, Yang and Ng, See-Kiong and Bing, Lidong and Lee, Roy Ka-Wei},
  journal={arXiv preprint arXiv:2406.17294},
  year={2024},
  url={https://arxiv.org/abs/2406.17294}
}

@article{zhuang2024math,
  title={Math-puma: Progressive upward multimodal alignment to enhance mathematical reasoning},
  author={Zhuang, Wenwen and Huang, Xin and Zhang, Xiantao and Zeng, Jin},
  journal={arXiv preprint arXiv:2408.08640},
  year={2024},
  url={https://arxiv.org/abs/2408.08640}
}

@inproceedings{han24infimm,
  title={InfiMM-WebMath-40B: Advancing Multimodal Pre-Training for Enhanced Mathematical Reasoning},
  author={Han, Xiaotian and Jian, Yiren and Hu, Xuefeng and Liu, Haogeng and Wang, Yiqi and Fan, Qihang and Ai, Yuang and Huang, Huaibo and He, Ran and Yang, Zhenheng and others},
  booktitle={The 4th Workshop on Mathematical Reasoning and AI at NeurIPS'24},
  year={2024},
  url={https://arxiv.org/abs/2409.12568}
}

@article{peng2024multimath,
  title={Multimath: Bridging visual and mathematical reasoning for large language models},
  author={Peng, Shuai and Fu, Di and Gao, Liangcai and Zhong, Xiuqin and Fu, Hongguang and Tang, Zhi},
  journal={arXiv preprint arXiv:2409.00147},
  year={2024},
  url={https://arxiv.org/abs/2409.00147}
}

@misc{openai2024gpt4omini,
  title        = {GPT-4o mini: advancing cost-efficient intelligence},
  author       = {{OpenAI}},
  year         = {2024},
  howpublished = {OpenAI product announcement},
  url          = {https://openai.com/index/gpt-4o-mini-advancing-cost-efficient-intelligence/},
}

@misc{bai2025qwen25vl,
      title={Qwen2.5-VL Technical Report}, 
      author={Shuai Bai and Keqin Chen and Xuejing Liu and Jialin Wang and Wenbin Ge and Sibo Song and Kai Dang and Peng Wang and Shijie Wang and Jun Tang and Humen Zhong and Yuanzhi Zhu and Mingkun Yang and Zhaohai Li and Jianqiang Wan and Pengfei Wang and Wei Ding and Zheren Fu and Yiheng Xu and Jiabo Ye and Xi Zhang and Tianbao Xie and Zesen Cheng and Hang Zhang and Zhibo Yang and Haiyang Xu and Junyang Lin},
      year={2025},
      eprint={2502.13923},
      archivePrefix={arXiv},
      primaryClass={cs.CV},
      url={https://arxiv.org/abs/2502.13923}, 
}

@article{bai2025qwen3vl,
  title         = {Qwen3-VL Technical Report},
  author        = {Bai, Shuai and Cai, Yuxuan and Chen, Ruizhe and Chen, Keqin and Chen, Xionghui and Cheng, Zesen and others},
  journal       = {arXiv preprint arXiv:2511.21631},
  year          = {2025},
  eprint        = {2511.21631},
  archivePrefix = {arXiv},
  primaryClass  = {cs.CV},
  url           = {https://arxiv.org/abs/2511.21631}
}

@article{wang2025internvl35,
  title         = {InternVL3.5: Advancing Open-Source Multimodal Models in Versatility, Reasoning, and Efficiency},
  author        = {Wang, Weiyun and Gao, Zhangwei and Gu, Lixin and Pu, Hengjun and Cui, Long and Wei, Xingguang and others},
  journal       = {arXiv preprint arXiv:2508.18265},
  year          = {2025},
  eprint        = {2508.18265},
  archivePrefix = {arXiv},
  primaryClass  = {cs.CV},
  url           = {https://arxiv.org/abs/2508.18265}
}

@misc{liu2026gdpogrouprewarddecouplednormalization,
      title={GDPO: Group reward-Decoupled Normalization Policy Optimization for Multi-reward RL Optimization}, 
      author={Shih-Yang Liu and Xin Dong and Ximing Lu and Shizhe Diao and Peter Belcak and Mingjie Liu and Min-Hung Chen and Hongxu Yin and Yu-Chiang Frank Wang and Kwang-Ting Cheng and Yejin Choi and Jan Kautz and Pavlo Molchanov},
      year={2026},
      eprint={2601.05242},
      archivePrefix={arXiv},
      primaryClass={cs.CL},
      url={https://arxiv.org/abs/2601.05242}, 
}

@article{chen2025whereMllms,
  title     = {Where {MLLMs} Attend and What They Rely On: Explaining Autoregressive Token Generation},
  author    = {Chen, Ruoyu and Guo, Xiaoqing and Liu, Kangwei and Liang, Siyuan and Liu, Shiming and Zhang, Qunli and Wang, Laiyuan and Zhang, Hua and Cao, Xiaochun},
  journal   = {arXiv preprint arXiv:2509.22496},
  year      = {2025},
  url       = {https://arxiv.org/abs/2509.22496}
}

@misc{chen2024expandingperformanceboundariesopensource,
      title={Expanding Performance Boundaries of Open-Source Multimodal Models with Model, Data, and Test-Time Scaling}, 
      author={Zhe Chen and Weiyun Wang and Yue Cao and Yangzhou Liu and Zhangwei Gao and Erfei Cui and Jinguo Zhu and Shenglong Ye and Hao Tian and Zhaoyang Liu and Lixin Gu and Xuehui Wang and Qingyun Li and Yiming Ren and Zixuan Chen and Jiapeng Luo and Jiahao Wang and Tan Jiang and Bo Wang and Conghui He and Botian Shi and Xingcheng Zhang and Han Lv and Yi Wang and Wenqi Shao and Pei Chu and Zhongying Tu and Tong He and Zhiyong Wu and Huipeng Deng and Jiaye Ge and Kai Chen and Kaipeng Zhang and Limin Wang and Min Dou and Lewei Lu and Xizhou Zhu and Tong Lu and Dahua Lin and Yu Qiao and Jifeng Dai and Wenhai Wang},
      year={2024},
      eprint={2412.05271},
      archivePrefix={arXiv},
      primaryClass={cs.CV},
      url={https://arxiv.org/abs/2412.05271}, 
}

@article{wang2024qwen2,
  title   = {Qwen2-VL: Enhancing Vision-Language Model's Perception of the World at Any Resolution},
  author  = {Wang, Peng and Bai, Shuai and Tan, Sinan and Wang, Shijie and Fan, Zhihao and Bai, Jinze and Chen, Keqin and Liu, Xuejing and Wang, Jialin and Ge, Wenbin and others},
  journal = {arXiv preprint arXiv:2409.12191},
  year    = {2024},
  url     = {https://arxiv.org/abs/2409.12191}
}

@misc{shi2025mathcanvasintrinsicvisualchainofthought,
  title         = {MathCanvas: Intrinsic Visual Chain-of-Thought for Multimodal Mathematical Reasoning},
  author        = {Shi, Weikang and Yu, Aldrich and Fang, Rongyao and Ren, Houxing and Wang, Ke and Zhou, Aojun and Tian, Changyao and Fu, Xinyu and Hu, Yuxuan and Lu, Zimu and Huang, Linjiang and Liu, Si and Liu, Rui and Li, Hongsheng},
  year          = {2025},
  eprint        = {2510.14958},
  archivePrefix = {arXiv},
  primaryClass  = {cs.CV},
  doi           = {10.48550/arXiv.2510.14958},
  url           = {https://arxiv.org/abs/2510.14958}
}

@inproceedings{lu2021intergps,
  title     = {Inter-GPS: Interpretable Geometry Problem Solving with Formal Language and Symbolic Reasoning},
  author    = {Lu, Pan and Gong, Ran and Jiang, Shibiao and Qiu, Liang and Huang, Siyuan and Liang, Xiaodan and Zhu, Song-Chun},
  booktitle = {Proceedings of the 59th Annual Meeting of the Association for Computational Linguistics and the 11th International Joint Conference on Natural Language Processing (Volume 1: Long Papers)},
  pages     = {6774--6786},
  year      = {2021},
  month     = aug,
  publisher = {Association for Computational Linguistics},
  url       = {https://aclanthology.org/2021.acl-long.528/}
}

@misc{kazemi2023geomverse,
  title         = {GeomVerse: A Systematic Evaluation of Large Models for Geometric Reasoning},
  author        = {Kazemi, Mehran and Alvari, Hamidreza and Anand, Ankit and Wu, Jialin and Chen, Xi and Soricut, Radu},
  year          = {2023},
  eprint        = {2312.12241},
  archivePrefix = {arXiv},
  primaryClass  = {cs.CV},
  doi           = {10.48550/arXiv.2312.12241},
  url           = {https://arxiv.org/abs/2312.12241}
}

@inproceedings{cao2022mgeo,
  title     = {An Augmented Benchmark Dataset for Geometric Question Answering through Dual Parallel Text Encoding},
  author    = {Cao, Jie and Xiao, Jing},
  booktitle = {Proceedings of the 29th International Conference on Computational Linguistics},
  pages     = {1511--1520},
  year      = {2022},
  month     = oct,
  publisher = {International Committee on Computational Linguistics},
  url       = {https://aclanthology.org/2022.coling-1.130/},
  doi       = {10.18653/v1/2022.coling-1.130}
}

@inproceedings{seo2015solving,
  title     = {Solving Geometry Problems: Combining Text and Diagram Interpretation},
  author    = {Seo, Minjoon and Hajishirzi, Hannaneh and Farhadi, Ali and Etzioni, Oren and Malcolm, Clint},
  booktitle = {Proceedings of the 2015 Conference on Empirical Methods in Natural Language Processing},
  pages     = {1466--1476},
  year      = {2015},
  month     = sep,
  publisher = {Association for Computational Linguistics},
  url       = {https://aclanthology.org/D15-1171/},
  doi       = {10.18653/v1/D15-1171}
}

@inproceedings{chen2022unigeo,
  title     = {UniGeo: Unifying Geometry Logical Reasoning via Reformulating Mathematical Expression},
  author    = {Chen, Jiaqi and Li, Tong and Qin, Jinghui and Lu, Pan and Lin, Liang and Chen, Chongyu and Liang, Xiaodan},
  booktitle = {Proceedings of the 2022 Conference on Empirical Methods in Natural Language Processing},
  pages     = {3313--3323},
  year      = {2022},
  month     = dec,
  publisher = {Association for Computational Linguistics},
  url       = {https://aclanthology.org/2022.emnlp-main.218/},
  doi       = {10.18653/v1/2022.emnlp-main.218}
}

@article{team2023gemini,
  title         = {Gemini: A Family of Highly Capable Multimodal Models},
  author        = {{Gemini Team} and Anil, Rohan and Borgeaud, Sebastian and Alayrac, Jean-Baptiste and Yu, Jiahui and Soricut, Radu and Schalkwyk, Johan and Dai, Andrew M. and Hauth, Anja and Millican, Katie and others},
  journal       = {arXiv preprint arXiv:2312.11805},
  year          = {2023},
  eprint        = {2312.11805},
  archivePrefix = {arXiv},
  doi           = {10.48550/arXiv.2312.11805},
  url           = {https://arxiv.org/abs/2312.11805}
}
